\documentclass[10pt,journal,compsoc]{IEEEtran}
\usepackage{amsmath,amsfonts}
\usepackage{array}
\usepackage{textcomp}
\usepackage{stfloats}
\usepackage{url}
\usepackage{verbatim}
\usepackage{graphicx}

\usepackage{eqparbox}
\usepackage{algorithm}
\usepackage{algorithmicx}
\usepackage{algpseudocode}
\usepackage{subfigure}
\usepackage{wrapfig}

\usepackage{xcolor}         % colors
\def\BibTeX{{\rm B\kern-.05em{\sc i\kern-.025em b}\kern-.08em
    T\kern-.1667em\lower.7ex\hbox{E}\kern-.125emX}}
\usepackage{balance}
\begin{document}
\title{Deep Probabilistic Graph Matching}
\author{He Liu,
        Tao Wang,
        Yidong Li,
        Congyan Lang,
        Songhe Feng,
        and Haibin Ling
\thanks{H. Liu is a Ph. D. student of the School of Computer and Information Technology, Beijing jiaotong University, Beijing 100044, China. e-mail: liuhe1996@bjtu.edu.cn.}% <-this % stops a space
\thanks{T. Wang, Y. Li, C. Lang and S. Feng are with the School of Computer and Information Technology, Beijing jiaotong University, Beijing 100044, China. e-mail: \{twang, ydli, cylang, shfeng\}@bjtu.edu.cn}% <-this % stops a space
\thanks{H. Ling is with the Department of Computer Science, Stony Brook University, Stony Brook, NY 11794-2424, USA. e-mail:\{haibin.ling@gmail.com\}}%
}

\markboth{Under Review}%
{He \MakeLowercase{\textit{et al.}}: Deep Probabilistic Graph Matching}

\maketitle

\begin{abstract}
Most previous learning-based graph matching algorithms solve the \textit{quadratic assignment problem} (QAP) by dropping one or more of the matching constraints and adopting a relaxed assignment solver to obtain sub-optimal correspondences. Such relaxation may actually weaken the original graph matching problem, and in turn hurt the matching performance.
In this paper we propose a deep learning-based graph matching framework that works for the original QAP without compromising on the matching constraints. In particular, we design an affinity-assignment prediction network to jointly learn the pairwise affinity and estimate the node assignments, and we then develop a differentiable solver inspired by the probabilistic perspective of the pairwise affinities. Aiming to obtain better matching results, the probabilistic solver refines the estimated assignments in an iterative manner to impose both discrete and one-to-one matching constraints.
The proposed method is evaluated on three popularly tested benchmarks (Pascal VOC, Willow Object and SPair-71k), and it outperforms all previous state-of-the-arts on all benchmarks.
\end{abstract}


\section{Introduction}\label{sec:intro}
\IEEEPARstart{G}{enerally} speaking, the problem of graph matching aims to find the optimal vertex correspondences between given graphs under structure constraints of keeping the edge consistency. It has been widely used in many applications such as object tracking~\cite{GM_visualTrack,Gracker}, person re-identification~\cite{reid}, point correspondence\cite{GMSP}, \textit{etc}.
Graph matching is in general NP-hard due to its combinatorial nature, and it is thus hard to search a global optimal solution for graphs with large sizes. Therefore, many approximate approaches~\cite{RWGM,BGM,PBGM,Graduated_Assignment,Spec_Tech,GNCCP,ABPF,PFGM,graph_hypergraph,Fact_GM} have been devoted to seeking acceptable suboptimal solutions by relaxing the quadratic assignment problem to a simpler form.

Aiming to improve the matching accuracy in real-world matching tasks, some early efforts~\cite{Learning_GM,LG2M} have been devoted to learning reasonable affinity measures or adaptive graph representations using the machine learning strategies. However, the improvements are limited because of the shallow parameter settings, and thus still insufficient to handle various challenges.

Recently, approaches~\cite{DGM,PCA,LGM,qcDGM,GLUE,BBGM,DGM_consensus,Hungarian_Attention} based on deep neural networks have attracted much research attention due to the ability to learn representative embeddings of nodes and/or edges. It is common for these approaches to embed a differentiable solver for the combinatorial optimization problem into an end-to-end learning framework. Different from the well-designed combinatorial solvers in previous learning-free graph matching methods, these learning-based methods usually acquire node correspondences in a more straightforward way by relaxing one or more of the quadratic constraints, discrete constraints and one-to-one matching constraints. Examples include~\cite{DGM_consensus,GLUE,PCA} that relax the quadratic assignment problem to a node-wise assignment problem and adopt the Sinkhorn algorithm~\cite{Sinkhorn_network} for the optimization of linear assignment, \cite{DGM} that works directly on the learned pairwise affinities using a spectral matching algorithm~\cite{Spec_Tech} but drops both discrete and one-to-one matching constraints, and~\cite{BBGM} that relaxes the graph matching problem based on Lagrangian decomposition and employs dual block coordinate ascent implementations for optimization. Despite remarkable performance gained, these relaxations of the original problem may cause potential limitations on matching performance.

To address the issues mentioned above, we propose a novel \emph{deep probabilistic graph matching} (DPGM) algorithm, which works directly on the learned pairwise affinities and imposes both discrete and one-to-one matching constraints on the matching solutions. Firstly, we design an \emph{affinity-assignment prediction network} (AA-predictor) to jointly learn the pairwise affinities and the node assignments. Specifically, the AA-predictor performs graph propagation using several delicately designed convolution operators on a constructed \emph{affinity-assignment graph} (AA-graph), in which each node corresponds to a candidate match and each edge relates to the pairwise affinity between two matches. Subsequently, the learned affinities and estimated assignments of the AA-predictor are passed as inputs to a \emph{differentiable probabilistic solver}, which refines the estimated assignments in an iterative manner and imposes both discrete and one-to-one matching constraints in a probabilistic way. Finally, the balanced entropy loss between the output matching solutions and ground-truth matches is employed as the supervision signal to guide the training of our framework.

For evaluating the proposed DPGM algorithm, we report its matching performance on three public benchmarks, namely Willow Objects~\cite{LG2M}, Pascal VOC Keypoints~\cite{VOC} and SPair-71k~\cite{SPair}, in comparison with several state-of-the-art graph matching approaches.
In experiments our method outperforms all compared methods on all three benchmarks, which illustrates the effectiveness and adaptability of the proposed method in different scenarios.
We will release our code publicly available once the paper is accepted.

In summary, with the proposed learning framework for graph matching, this paper makes contribution in three-fold:
\begin{enumerate}
    \item    the proposed differentiable probabilistic solver works directly on the learned pairwise affinities in a probabilistic way, which is expected to be more effective by avoiding the compromise on the matching constraints;
    \item unlike most previous graph matching methods that use a trivial guess of an initial assignment for the combinatorial solver, ours learns jointly the pairwise affinities and the initial assignment, which remarkably benefits the final solution (see Tables~\ref{table:ablation},~\ref{table:abl_spair},~\ref{table:abl_willow}); and
    \item  extensive experiments are conducted on three popular benchmarks, in comparison with many state-of-the-art methods, to validate the effectiveness of the proposed method.
\end{enumerate}

\section{Related Works}

Graph matching has been investigated for decades and many algorithms have been proposed. In this section we review recent learning-based studies or those closely related to ours, and leave general graph matching research to three comprehensive surveys~\cite{Conte2004,Foggia2014,survey_yan}.

Aiming to cooperate with the data derived from real-world matching tasks, many learning-based graph matching algorithms, including unsupervised~\cite{unsup_GM}, semi-supervised~\cite{semisup_GM} and supervised ones~\cite{Learning_GM,smooth_optim}, have been proposed to learn the parameters of affinity measure to replace the handcrafted affinity metric. In addition, instead of learning the affinity measure, Cho \textit{et al.}~\cite{LG2M} propose a learnable framework to parameterize the graph model and generate reasonable structural attributes for visual object matching. However, these methods employ simple and shallow parametric models to control geometric affinities between pairs of matches, and the promotion to the matching accuracy is still limited.

With the growing interest in utilizing deep neural network for structured data~\cite{TIP1, TIP2}, learning graph matching with \emph{graph neural network} (GNN) has attracted much research attention. A classic way is to learn representative node and/or edge embeddings via graph neural networks and then relax the graph matching problem to the linear assignment problem.
Nowak~\textit{et al.}~\cite{NowakVBB18} introduce a Siamese GNN encoder to produce a normalized node embedding for each graph to be matched, and then predict a matching by minimizing the cosine distance between matching pairs of embeddings dictated by the target permutation.
Wang~\textit{et al.}~\cite{PCA} employ the \textit{graph convolutional network} (GCN) framework~\cite{KipfW17} to produce node embeddings by aggregating graph structure information, and adopt the Sinkhorn network~\cite{Sinkhorn_network} as the combinatorial solver for the relaxed linear assignment problem.
SuperGlue~\cite{GLUE} is designed for generating discriminative node representations using intra-graph and cross-graph attention, where spatial relationships and visual information are jointly taken into considerations during the node embedding process. In addition, to drop the outlier candidate matches that usually occur in real-world matching tasks, this framework augments each node set with a dustbin so that the unmatched nodes are explicitly assigned to it.
To further improve the robustness of learnable affinities, Yu~\textit{et al.}~\cite{Hungarian_Attention} propose a node and edge embedding strategy that simulates the multi-head strategy in attention models, enabling the information in each channel to be merged independently.
Fey~\textit{et al.}~\cite{DGM_consensus} propose to start from an initial ranking of soft correspondences between nodes, and iteratively refine the solution by synchronous message passing networks to reach neighborhood consensus between node pairs without any optimization inference.

Different from the above mentioned methods that focus mainly on the learning of node and/or edge embeddings and relax the quadratic matching constraints,
several recently proposed methods work directly on pairwise affinities and embed differentiable solvers for quadratic optimization.
Zanfir~\textit{et al.}~\cite{DGM} formulate graph matching as a quadratic assignment problem under both unary and pairwise affinities, and adopt a spectral matching algorithm~\cite{Spec_Tech} as the combinatorial solver that drops both discrete and one-to-one matching constraints in optimization.
Rol{\'{\i}}nek~\textit{et al.}~\cite{BBGM} relax the graph matching problem based on Lagrangian decomposition, which is solved by
embedding BlackBox implementations of a heavily optimized solver~\cite{SwobodaRAKS17} based on dual block coordinate ascent.
Reformulating graph matching as Koopmans-Beckmann's QAP~\cite{LoiolaANHQ07} to minimize the adjacency discrepancy of graphs to be matched, Gao \textit{et al.}~\cite{qcDGM} adopt the Frank-Wolf algorithm~\cite{Frank&Wolfe56} to obtain approximate solutions.
Besides, Wang~\textit{et al.}~\cite{LGM} integrate learning of affinities and solving for combinatorial optimization into a unified node labeling pipeline, where the quadratic assignment problem of graph matching has been transformed to the binary classification problem of finding the positive node in a constructed assignment graph.

\begin{figure*}[htb]
\begin{center}
\includegraphics[width=\linewidth,height=4.8cm, trim=0 100 30 0,clip]{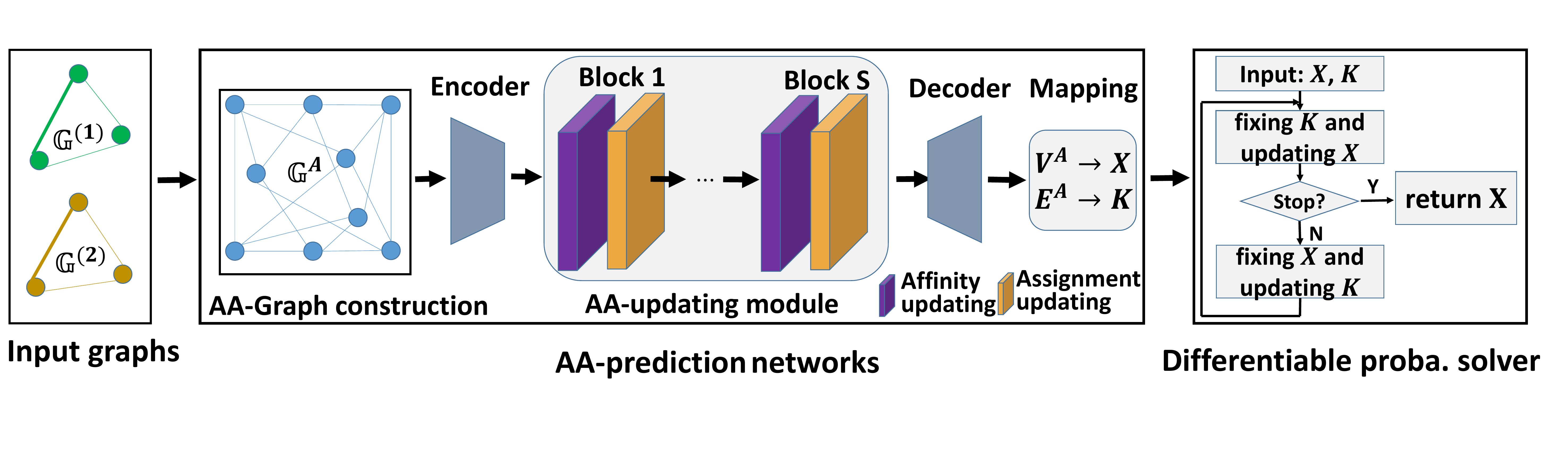}
\end{center}
\caption{The overview pipeline of the proposed DPGM framework. Taking the two graphs $\mathbb{G}^{(1)}$ and $\mathbb{G}^{(2)}$ to be matched as input, the AA-predictor firstly generates an affinity-assignment graph (AA-graph) $\mathbb{G}^A$, where the candidate matches and pairwise affinities are modeled as nodes and edges respectively. Subsequently, the graph state is iteratively updated by the AA-updating module to form structured representations, and it is finally read out by the decoder that maps representations of nodes and edges to entries of the assignment matrix and the pairwise affinity matrix respectively. Finally, the differentiable probabilistic solver starts from the assignments estimated by the AA-predictor, and refines them alternatively by solving QAP in a probabilistic way to obtain better matching solutions. }
\label{pipeline}
\end{figure*}

\section{Graph Matching}
\subsection{Problem formulation}
In graph theory, an attributed graph of $n$ nodes can be represented by $\mathbb{G}=\{\mathbb{V},\mathbb{E},\mathcal{V},\mathcal{E}\}$, where $\mathbb{V}=\{v_1,...,v_n\}$ and $\mathbb{E}\subseteq\mathbb{V}\times{\mathbb{V}}$ denote respectively the node set and edge set, and $\mathcal{V}=\{\textbf{v}_i|\textbf{v}_i\in \mathbb{R}^{d_V},i=1,2,...,n\}$ and $\mathcal{E}=\{\textbf{e}_i|\textbf{e}_i\in \mathbb{R}^{d_E},i=1,2,...,|\mathcal{E}|\}$ the node attribute set and edge attribute set, respectively. The node relations in the graph can be represented by a symmetric adjacency matrix $A\in \{0,1\}^{n\times{n}}$, where $A_{ij}=1$ if and only if there is an edge between nodes $v_i$ and $v_j$.

Given two graphs $\mathbb{G}^{(i)}=\{\mathbb{V}^{(i)},\mathbb{E}^{(i)},\mathcal{V}^{(i)},\mathcal{E}^{(i)}\}$ of size $n$, $i=1,2$, graph matching aims to find node correspondence $X\in \{0,1\}^{n\times{n}}$ between $\mathbb{G}^{(1)}$ and $\mathbb{G}^{(2)}$ that maximizes the global consistency designed as
\begin{equation}
\label{global_consistency}
\mathcal{H}=\sum_{i,a}c_{ia}X_{ia}+\sum_{i,j,a,b}d_{ia,jb}X_{ia}X_{jb},
\end{equation}
where $c_{ia}$ measures the similarity between node $v_{i}^{(1)}$ in $\mathbb{G}^{(1)}$ and node $v_{a}^{(2)}$ in $\mathbb{G}^{(2)}$, while $d_{ia,jb}$ measures the agreement between edge $(v_{i}^{(1)},v_{j}^{(1)})$ in $\mathbb{G}^{(1)}$ and edge $(v_{a}^{(2)},v_{b}^{(2)})$ in $\mathbb{G}^{(2)}$. The correspondence matrix $X$ indicates the matching results, i.e., $X_{ia}=1$ if and only if $i^{th}$ node of $\mathbb{G}^{(1)}$ matches to $a^{th}$ node of $\mathbb{G}^{(2)}$. In addition, the distribute of $X$ is restricted under the one-to-one matching constraints: $X\textbf{1}_n=\textbf{1}_n$ and $X^T\textbf{1}_n=\textbf{1}_n$, where $\textbf{1}_n$ denotes a $n-$demension one-value vector.

Seeking the optimal solution maximizing the Eq.~\ref{global_consistency} can be reformulated as a quadratic assignment problem:
\begin{equation}\label{graph_matching}
\textbf{x}^\ast=\arg\max_\textbf{x} \textbf{x}^TK\textbf{x},
\end{equation}
where $\textbf{x}$ is the vectorization of $X$, and $K$ the affinity matrix encoding the node similarity and edge agreement at diagonal elements and off-diagonal elements, respectively. In more detail, the affinity matrix $K$ can be expressed as
\begin{equation}
\label{K_mat}
K_{ia,jb}=
\left\{
    \begin{array}{ll}
    c_{ia}         & {\rm if~~} i=j {\rm ~~and~~}  a=b,\\
    d_{ia,jb} &     {\rm else~if~~}  A_{ia}^{(1)}A_{jb}^{(2)}>0,\\
    0                   & {\rm otherwise.}
    \end{array}
\right.
\end{equation}

Note that, for simplicity we assume that the two given graphs to be matched have the same size $n$ in this work, yet this formulation of graph matching can be easily extended to general cases with different sizes by auxiliary strategies, such as adding dummy nodes.

Recently, many efforts have been devoted to improving graph matching accuracy with the combination of deep learning architecture and differentiable relaxation-based solvers. The graph matching problem has been relaxed to a linear assignment problem by learning the high-order node embeddings in~\cite{PCA,DGM_consensus,Hungarian_Attention}, and the Sinkhorn algorithm~\cite{sinkhorn} is applied as a combinatorial solver in these works. In~\cite{DGM}, the unary and pair-wise affinities are generated by learning deep node and edge representations, and the quadratic assignment problem is solved by a relaxation manner, \textit{i.e.}, spectral matching~\cite{Spec_Tech}. A more advanced method has been proposed by Rol{\'{\i}}nek~\textit{et al.}~\cite{BBGM} who propose using strong feature extraction with SplineCNN~\cite{splineCNN} and leverage the combinatorial solver based on dual block coordinate ascent for Lagrange decompositions~\cite{SwobodaRAKS17}. Despite the demonstrated power of deep networks in representation learning, the relaxation strategy applied to graph matching problem is a weakening of the quadratic assignment problem, which may hurt the performance of graph matching.

Unlike the deep graph matching methods discussed above, we treat the graph matching problem as a quadratic assignment problem faithfully without compromise on matching constraints, and solve the problem using a probabilistic optimization scheme that takes as input the estimated affinities and assignments, both of which are jointly learnt through a novel graph information propagation module, and refines the assignment solutions in an iterative manner.

\section{The Proposed Method}

Our model consists of two core parts: an \textit{affinity-assignment prediction network} (AA-predictor) and a \textit{differentiable probabilistic solver}. As illustrated in Fig.~\ref{pipeline}, the AA-predictor first models the pairwise affinity and candidate matches into a unified \textit{affinity-assignment graph} (AA-graph), and then performs graph propagation to form the structural representations for nodes and edges, both of which are decoded as the initial assignments and affinities, respectively. Subsequently, the differentiable probabilistic solver takes the estimated assignments and affinities as input, and refines them in a probabilistic manner to obtain the optimal matching solutions.

\subsection{Affinity-assignment prediction network}
As illustrated in Fig.~\ref{pipeline}, the AA-predictor takes two graphs to be matched as input, and learns pairwise affinities and candidate assignments through three components including the AA-graph construction, AA-updating module and AA-decoder.

\textbf{Affinity-assignment graph construction:}
Given two input graphs $\mathbb{G}^{(1)}$ and $\mathbb{G}^{(2)}$, we model the candidate matches and the pairwise affinities into a so-called \textit{affinity-assignment graph} (AA-graph) $\mathbb{G}^A=\{\mathbb{V}^A,\mathbb{E}^A,\mathcal{V}^A,\mathcal{E}^A\}$, where the candidate match between $v_i^{(1)} \in \mathbb{V}^{(1)}$ and $v_a^{(2)} \in \mathbb{V}^{(2)}$ is denoted as a node $v_{ia}^A \in \mathbb{V}^A$, and the pair-wise affinity between a pair of candidate matches $v_{ia}^A, v_{jb}^A \in \mathbb{V}^A$ is represented by the edge $e_{(ia,jb)}^A \in \mathbb{E}^A$ if and only if there are two edges $(v_i^{(1)}, v_j^{(1)})\in \mathbb{E}^{(1)}$ and $(v_a^{(2)}, v_b^{(2)})\in \mathbb{E}^{(2)}$.

\begin{figure}[h]
%\vspace{-6mm}
\centering
\includegraphics[width=9.5cm,height=4cm, trim=0 100 0 50,clip]{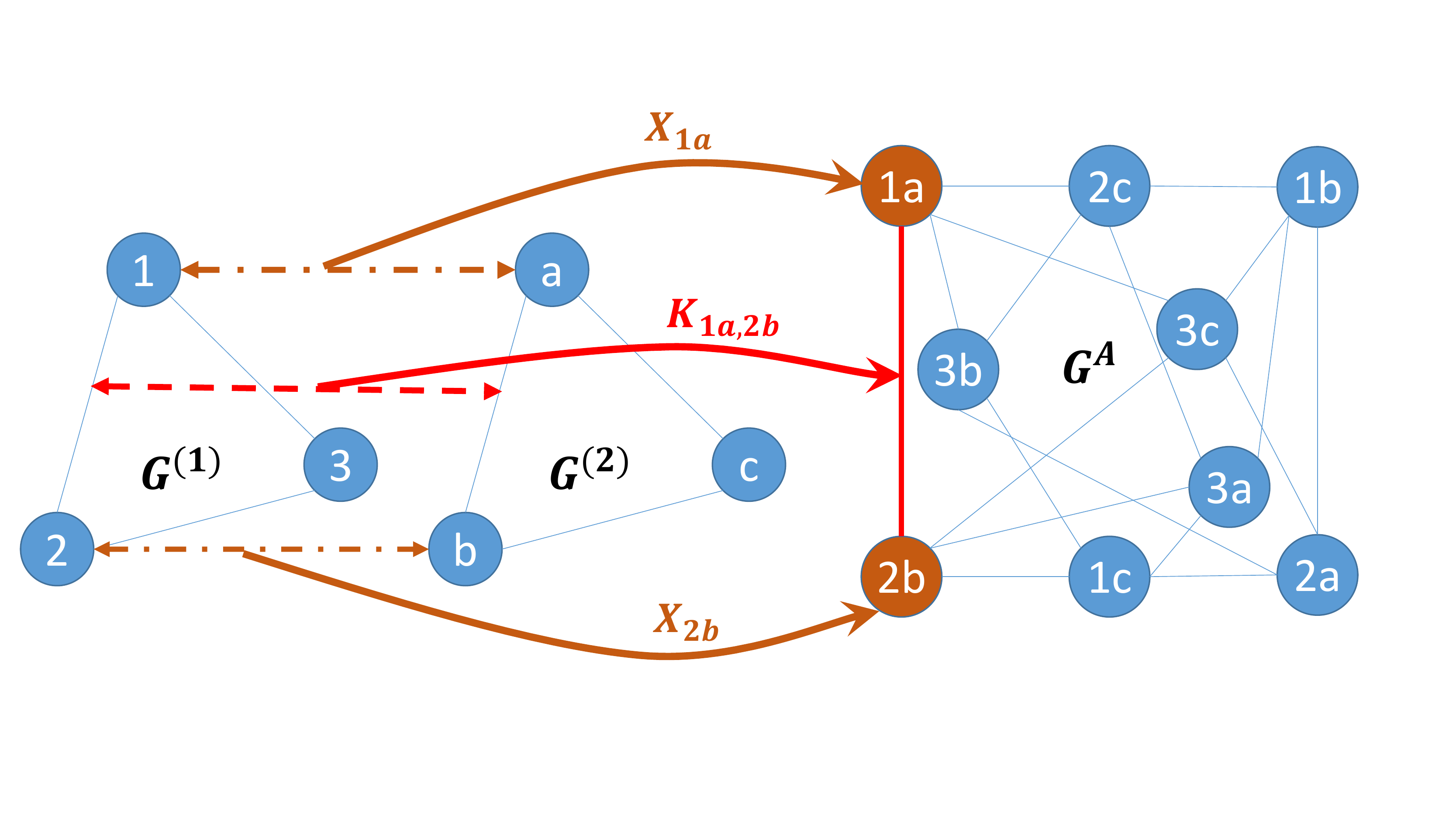}
%\vspace{-3mm}
\caption{An example of the AA-graph construction.
%With the two input graphs $\mathbb{G}^{(1)}$ and $\mathbb{G}^{(2)}$, the candidate matches and affinities are modeled into the nodes and edges in the constructed Affinity-assignment graph $\mathbb{G}^{A}$. For this illustration, the candidate matches $X_{1a}$ and $X_{2b}$ are modeled as the nodes $v_{1a}^A, v_{2b}^A \in \mathbb{V}^A$, and the affinity between them, i.e., $K_{1a,2b}$, is encoded into the edge $e_{(1a,2b)}^A \in \mathbb{E}^A$.
}
\label{subfig:AAgraph}
\end{figure}

Fig.~\ref{subfig:AAgraph} illustrates an example of constructing the AA-graph where the candidate matches $X_{1a}$ and $X_{2b}$ are encoded as the nodes $v_{1a}^A$ and $v_{2b}^A$ in $\mathbb{V}^A$ respectively, and the affinity between them (i.e., $K_{1a,2b}$) is modeled into the edge $(v_{1a}^A,v_{2b}^A) \in \mathbb{E}^A$.
Different from the method proposed in~\cite{LGM} that utilizes only geometric cues to form the attributes of nodes and edges, we combine the visual features and geometric information to generate the initial attributes on AA-graph. Specifically, we take the visual features extracted from images by CNNs~\cite{CNN} as the raw node attributes, and concatenate the point coordinates of two nodes associated with the same edge to form the edge attributes as %. Therefore, the initial features on node $v_{ia}^A$ and edge $e_{(ia,jb)}^A$ are expressed as
\begin{equation}\label{raw_feature}
    \begin{aligned}
    \textbf{v}_{ia}^A & =[\textbf{f}_i^{(1)}; \textbf{f}_a^{(2)}];  \\
    \textbf{e}_{(ia,jb)}^A & = [\textbf{p}_{i}^{(1)}; \textbf{p}_{j}^{(1)}; \textbf{p}_{a}^{(2)}; \textbf{p}_{b}^{(2)}],
    \end{aligned}
\end{equation}
where $\textbf{f}_i^{(1)}$ and $\textbf{p}_{i}^{(1)}$ denote the visual feature and coordinates of the $i^{th}$ keypoint in the $1^{st}$ image respectively, and $[\cdot;\ldots;\cdot]$ concatenates its input along the channel direction.

In this way, the affinity matrix and matches in Eq.~\ref{graph_matching} have been transformed into high-order representations in the unified AA-graph, and they are alternatively updated to form structured representations by the following AA-updating module.

%============================AA_graph========================
%============================AA_graph========================

\textbf{AA-updating module:}
This module performs graph propagation by extending the graph network block~\cite{LGM} module to fit our framework.
Specifically, this module firstly transforms the original node attributes and edge attributes of the input AA-graph into two latent embedding spaces, and then iteratively updates the affinities (edge attributes) and the assignments (node attributes) by stacking multiple affinity updating layers and assignment updating layers.

For the original AA-graph, the AA-updating module employs the encoder to transform the AA-graph state into latent embedding space as
\begin{equation}
\mathbb{G}^A \leftarrow Encoder(\mathbb{G}^A) = \{\mathbb{V}^A, \mathbb{E}^A, \rho_v(\mathcal{V}^A), \rho_e(\mathcal{E}^A)\},
\end{equation}
where $\rho^v$ and $\rho^e$ are the learnable transformation functions that map the original attributes of nodes and edges into the latent spaces with dimensions of $d_V$ and $d_E$, respectively. Furthermore, both $\rho_v$ and $\rho_e$ are designed as two Multi-Layer Perceptrons (MLPs), but with different parameters. After that, the AA-updating module performs graph propagation to form the structural representations for nodes and edges.

Since a pairwise affinity $\textbf{e}_{(ia,jb)}^A$ is directly related to its associated matches $\textbf{v}_{ia}^A$ and $\textbf{v}_{jb}^A$, we design the affinity updating layer as
\begin{equation}
    \begin{aligned}
        \bar{\textbf{e}}_{(ia,jb)}^A & = (M_1 \textbf{v}_{ia}^A)\odot(M_2 \textbf{v}_{jb}^A); \\
        \textbf{e}_{(ia,jb)}^A & \leftarrow \tau([\textbf{e}_{(ia,jb)}^A; \bar{\textbf{e}}_{(ia,jb)}^A]),
    \end{aligned}
\end{equation}
where $M_1,M_2 \in \mathbb{R}^{d_V\times{d_V}}$ are the learnable similarity coefficients, the operator $\odot$ denotes the element-wise multiplication of two vectors, and $\tau$ is the update function designed as a learnable MLP that maps the concatenated features to the dimension of $d_E$.

For a candidate assignment $v_{ia}^A$, we wish to determine its reliability based on its associated pairwise affinities.
Therefore, in the assignment updating layer we update the attribute of each node by aggregating the information of its associated edges as

\begin{equation}
\textbf{v}_{ia}^A \leftarrow \kappa\Big(\big[\sum_{\textbf{e} \in \mathcal{E}^A_{ia}} \textbf{e};\textbf{v}_{ia}^A \big] \Big),
\end{equation}
where $\mathcal{E}^A_{ia}$ denotes the attribute set of the edges adjacent to node $v_{ia}^A$, and $\kappa$ the parameterized function designed as an MLP that maps the aggregated features to the dimension of $d_V$.

\textbf{AA-Decoder:}
The decoder can be interpreted as an inverse procedure of AA-graph construction, and it maps the structured representations of edges and nodes to scalar values in affinity matrix $K$ and assignment matrix $X$.
In particular, node attributes are transformed to scalar values in $(0,1)$ that indicate the probabilities of the corresponding matches, and edge attributes to scalar values in $(0,1)$ that denote the affinities of the corresponding pairs of candidate matches. %$[0,+\infty]$
The concrete transformation process in the decoder module can be expressed as
\begin{equation}\label{decoder}
    \begin{aligned}
        \textbf{X}_{ia} & = \mathrm{sigmoid}(\varphi_n(\textbf{v}_{ia}^A)); \\
        K_{ia,jb} & = \mathrm{sigmoid}(\varphi_e(\textbf{e}_{(ia,jb)}^A)),
    \end{aligned}
\end{equation}
where $\varphi_n$ and $\varphi_e$ are two learnable MLP-based update functions that transform the high-order features to the scalar values in $(-\infty,+\infty)$, and $\mathrm{sigmoid}(z)$ is the activation function formulated as $\mathrm{sigmoid}(z)=\frac{1}{1+e^{-z}}$,
% \begin{equation}
% \mathrm{sigmoid}(z)=\frac{1}{1+e^{-z}},
% \end{equation}
which maps its input into the values in $(0,1)$.

\subsection{Differentiable Probabilistic Solver}\label{prob_estimation}
Instead of taking the estimated assignments from AA-predictor as the final matching solution, we design a differentiable probabilistic solver to refine the estimated assignments by imposing both discrete and one-to-one matching constraints.

There are several studies~\cite{graph_hypergraph,PBGM,RWGM} dedicated to solving Eq.~\ref{graph_matching} from the probability perspective, which have exhibited state-of-the-art matching performance among the family of learning-free algorithms. In the probabilistic interpretation of graph matching, the assignment $X_{ia}$ is regarded as the probability that the node $v_{i}^{(1)} \in \mathbb{V}^{(1)}$ matches $v_{a}^{(2)} \in \mathbb{V}^{(2)}$, and the affinity $K_{ia,jb}$ is considered as the empirical estimation of the pairwise assignment probability such that $v_{i}^{(1)} \in \mathbb{V}^{(1)}$ matches $v_{a}^{(2)} \in \mathbb{V}^{(2)}$ and $v_{j}^{(1)} \in \mathbb{V}^{(1)}$ matches $v_{b}^{(2)} \in \mathbb{V}^{(2)}$.

Denoting $\mathcal{P}_X=\{P(X_{ia}):1\leq i,a\leq n\}$ in which $P(X_{ia})$ is the probability that the match between $v_{i}^{(1)} \in \mathbb{V}^{(1)}$ and $v_{a}^{(2)} \in \mathbb{V}^{(2)}$ is valid;
and $\mathcal{P}_{X,X}=\{P(X_{ia}|X_{jb}):1\leq i,a,j,b\leq n\}$ in which $P(X_{ia}|X_{jb})$ denotes the conditional assignment probability (\textit{i.e.}, the probability of the assignment $X_{ia}$ under the condition that the assignment $X_{jb}$ is valid). The graph matching problem of Eq.~\ref{graph_matching} that finding the optimal assignment results can be reformulated as
\begin{equation}\label{pro_GM}
\begin{split}
 &[\mathcal{P}^{\ast}_{X},\mathcal{P}^{\ast}_{X,X}] =\\
   & ~~~ \operatornamewithlimits{\arg \min}_{\mathcal{P}_{X}, \mathcal{P}_{X,X}} ~ \sum_{i,a}\Big( \sum_{j,b}{P(X_{ia}|X_{jb})P(X_{jb})} - P(X_{ia}) \Big)^2.
\end{split}
\end{equation}

It is common~\cite{graph_hypergraph,PBGM,RWGM} to solve such a probabilistic optimization through iterating a two-step optimization process: (1)
estimating the assignment probability $P(X_{ia})$ according to the current affinity distribution; and (2) refining conditional assignment probability $P(X_{ia}|X_{jb})$ upon the computed individual probability.

Thus inspired, we design a differentiable probabilistic solver that is summarized in Algorithm~\ref{algo_solver}.
For estimating the assignment probabilities, we firstly vectorize the predicted assignment matrix (Line 3), in which each element can be regarded as the probability of corresponding candidate match, i.e., $P(X_{jb})$. For the predicted affinities matrix, we consider elements in each column $K_{\cdot,jb}$ as the joint probabilities under the corresponding match $X_{jb}$ is valid, that is, $P(X_{ia}|X_{jb}): 1\leq i,a \leq n$. Thus, under the probabilistic explanations, the probabilistic solver updates the assignment probabilities using the current conditional probabilities using $\textbf{x}_{t+1} \leftarrow K_t \textbf{x}_t$ (Line 4). The updating for each element in $\textbf{x}$ can be interpreted as the following probabilistic formulation
%In the probabilistic view of graph matching, fixing the affinities and updating for assignments $\textbf{x}_{t+1} \leftarrow K_t \textbf{x}_t$ (Line 4) can be interpreted as a power iteration process~\cite{matrix} that is widely used to solve the spectral matching problem
\begin{equation}\label{step1}
  P_{t+1}(X_{ia})=\sum_{j,b}P_t(X_{ia}|X_{jb})P_t(X_{jb}).
\end{equation}
After that, the Sinkhorn algorithm~\cite{Sinkhorn_network} (Line 6) performs both row- and column-normalization on the assignment matrix, which guarantees the one-to-one matching constraints in a soft way.
In other words, the probabilities that each node $v_i^{(1)} \in \mathbb{V}^{(1)}$ matches all nodes $v_a^{(2)} \in \mathbb{V}^{(2)}$ should add up to 1, and vice versa.

During the refinement of the affinity matrix $K$ (Line 10, where $\oslash$ denotes element-wise division between two vectors), we adaptively increase the entries that correspond to valid assignments and weaken the ones that correspond to invalid assignments.
Specifically, if an updated assignment $\textbf{x}_{t+1}^{ia}$ between $v_i^{(1)} \in \mathbb{V}^{(1)}$ and $v_a^{(2)} \in \mathbb{V}^{(2)}$ is larger than the previous one $\textbf{x}_{t}^{ia}$, that is $\frac{\textbf{x}_{t+1}^{ia}}{\textbf{x}_{t}^{ia}}>1$, we think it is more likely to be a valid assignment. Otherwise, the assignment between $v_i^{(1)} \in \mathbb{V}^{(1)}$ and $v_a^{(2)} \in \mathbb{V}^{(2)}$ is more likely to be an invalid one. Thus, we multiply the incremental $\frac{\textbf{x}_{t+1}}{\textbf{x}_{t}}$ to augment or weaken the joint probabilities that it involves.
From the probabilistic perspective, this process actually equals to refining the conditional assignment probability $P(X_{ia}|X_{jb})$ according to the estimated $P(X_{ia})$ by
\begin{equation}\label{step2}
  P_{t+1}(X_{ia}|X_{jb})=P_{t}(X_{ia}|X_{jb})\frac{P_{t+1}(X_{ia})}{P_t(X_{ia})},
\end{equation}
where $t$ and $t+1$ indicate iteration steps.

%===========================================algorithm=======================================================
\begin{algorithm*}[t]
\caption{The differentiable probabilistic solver.}
\label{algo_solver}
\begin{algorithmic}[1]
\Require
$X_1$: initial assignment solution;
$K_1$: initial affinity matrix;
$S$: maximum of iteration;
$\eta$: predefined threshold.
\Ensure
optimal assignment solution $X^{*}$.
\vspace{1mm}
\State initialization: $X_t \leftarrow X_1$, $K_t \leftarrow K_1$;
%\State normalize the assignment matrix: $X^t=Sinkhorn(X^t)$
\For {$t=1$; $t\leq S$; $t++$ }
    \State $\textbf{x}_t \leftarrow \textrm{vec}(X_t)$; \Comment{vectorizing $X_t$}
%    \If{$t \mod 2 = 1$}
        \State $\textbf{x}_{t+1} \leftarrow K_t \textbf{x}_t$       \Comment {estimate the assignment probabilities}
        \State $X_{t+1} \leftarrow \textrm{reshape}(\textbf{x}_{t+1})$;          \Comment {reshape the probabilities to matrix form}
        \State $X_{t+1} \leftarrow \textrm{Sinkhorn}(X_{t+1})$;     \Comment {normalize the probabilities by Sinkhorn}
        \If {$\|\textbf{x}_{t+1} - \textbf{x}_t\|^2<\eta$}
            \State break;
        \EndIf
%    \Else
        \State $K_{t+1} \leftarrow K_t ((\textbf{x}_{t+1} \oslash \textbf{x}_{t} ) \textbf{1}^T)$; \Comment {update the conditional probabilities}
%    \EndIf
\EndFor
\State $X^*\leftarrow X_{t+1}$
\end{algorithmic}
\end{algorithm*}
%====================================algorithm done====================================================

To validate the sparsity of the estimated assignment matrix, we record its binary score at each iteration on Willow Object dataset~\cite{LG2M}, which is computed according to its $\ell_{2,1}$ norm as
\begin{equation}
  s=\frac{1}{2N}\big(\|X \|_{\ell_{2,1}} + \|X^\top \|_{\ell_{2,1}}\big),
\end{equation}
where $N$ is the number of nodes of the graphs to be matched.

As illustrated in Fig.~\ref{subfig:sparse}, the binary scores $s$ of the estimated assignments are gradually improved with the iteration increasing.
It approaches $1$ after several iterations, meaning the assignment matrix is nearly discrete under the premise that the one-to-one matching constraints are guaranteed by the Sinkhorn algorithm.
That is to say, the proposed solver is able to impose not only the one-to-one constraints but also the discrete constraints, % in the training step,
which are usually considered only in the inferring step in most previous end-to-end learning frameworks~\cite{DGM,PCA,LGM}.

\subsection{Loss Function}
Similar to~\cite{LGM,PCA}, we also utilize the difference between the predicted assignments and groundtruth node-to-node correspondences as the supervision signal to guide the training process. Specifically, given the groundtruth correspondences $X^{gt}$ and estimated solutions $X$, we first reshape them to the vector form $\textbf{x}^{gt}$ and $\textbf{x}$ respectively, and measure the difference between them using the balanced cross entropy loss
\begin{equation}\label{loss}
  \mathcal{L}= -\sum_{i=1}^{n^2} \Big[ w\ \textbf{x}_i^{gt} \log(\textbf{x}_i) \ + \ (1-w) (1-\textbf{x}_i^{gt}) \log(1-\textbf{x}_i) \Big],
\end{equation}
where $w$ is a hyper-parameter that balances the loss to avoid the dominance of the negative candidate matches during training.

\section{Experiments}\label{sec:exp}

To validate the effectiveness of our framework, we evaluate it on three public visual keypoint matching benchmarks, named Willow Object dataset~\cite{LG2M}, Pascal VOC Keypoints~\cite{VOC} and SPair-71k~\cite{SPair}, in comparison with ten state-of-the-art learning-based graph matching approaches including GMN~\cite{DGM}, PCA~\cite{PCA}, IPCA~\cite{PCA-PAMI}, LGM~\cite{LGM}, qc-DGM~\cite{qcDGM}, DGMC~\cite{DGM_consensus}, CIE~\cite{Hungarian_Attention}, BBGM~\cite{BBGM}, NGM~\cite{NGM} and NGMv2~\cite{NGM}. Among these baseline methods, IPCA~\cite{PCA-PAMI} is the upgraded version of PCA~\cite{PCA} by iterating the cross-graph updating for node attributes, and NGMv2~\cite{NGM} is the extension of NGM~\cite{NGM} with the refinement for initial graph features by SplineCNN networks~\cite{splineCNN}.
Although it has been demonstrated that adaptively adjusting graph structure can also achieve the state-of-the-art performance~\cite{DLGM}, in this paper we only focus on the works that estimate the matching solutions with the combination of deep learning framework and differentiable solvers under fixed graph topology. Therefore, the recently proposed method~\cite{DLGM} that adaptively generates latent graph topology for graph matching has been excluded from baseline methods.

%========================sparse==========================
\begin{figure}[!thb]
\centering
%\vspace{-7mm}
\includegraphics[width=8.5cm,height=5cm]{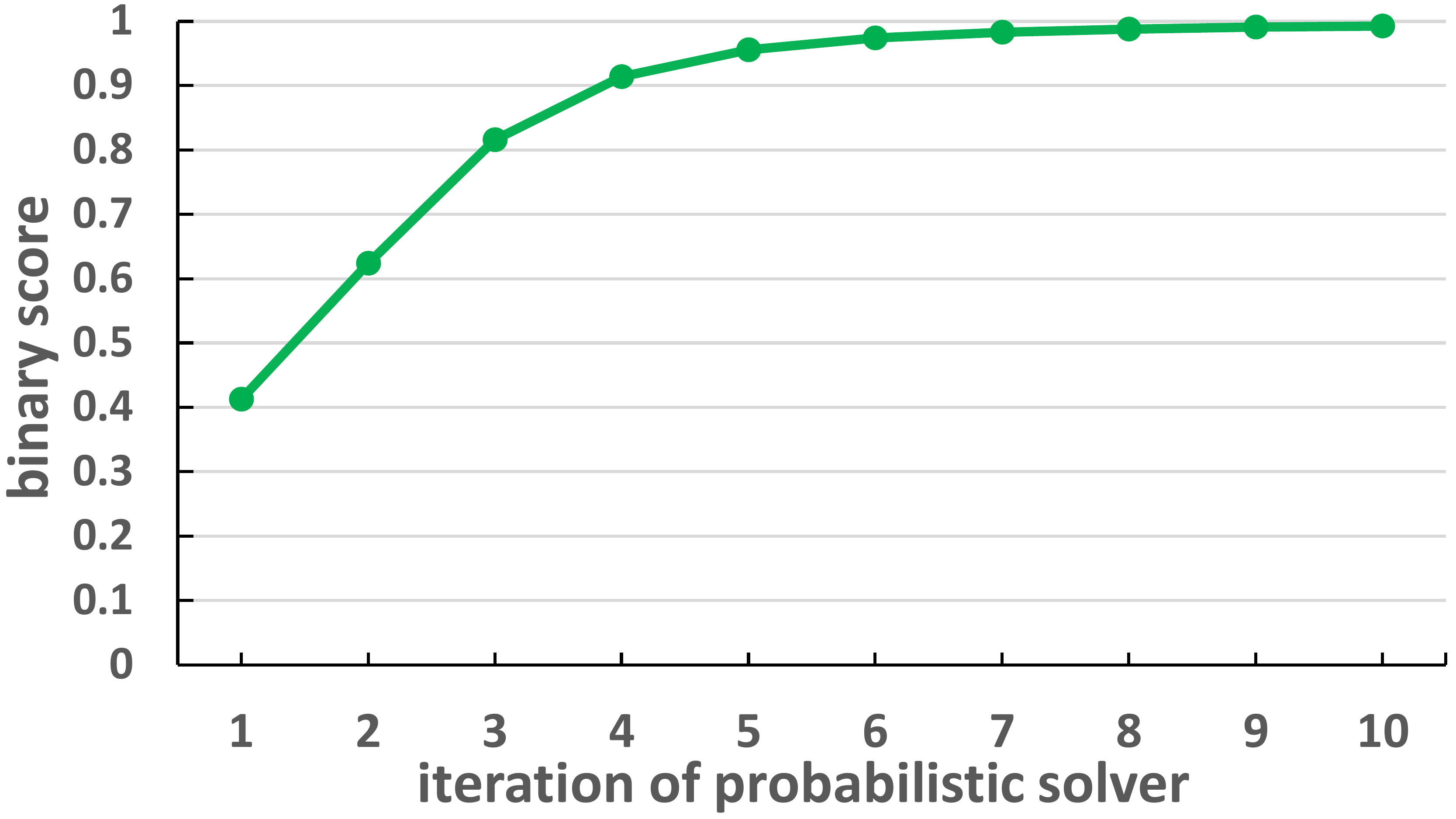}
%\vspace{-5mm}
\caption{The illustration for convergence of the proposed probabilistic solver.}
%\vspace{-5mm}
\label{subfig:sparse}
\end{figure}
%========================sparse==========================

In experiments, we follow~\cite{BBGM} to extract the visual features using the VGG16 backbone~\cite{VGG16} as the raw keypoint features. The encoder transforms both the initial node features and edge features on AA-graph to the 128-dimension latent space for the graph matching tasks on the Pascal VOC~\cite{VOC} and SPair-71k~\cite{SPair} datasets. In other words, both $d_V$ and $d_E$ are set as 128. For the much easier Willow~\cite{LG2M} dataset, the encoder transforms the AA-graph features to 32-dimension latent space (i.e., $d_V=d_E=32$) to alleviate the over-fitting during training.
For gradient back propagation in model learning, the gradients of the solution would be very small when the solution is close to $\{0,1\}$ after many iterations of the probabilistic solver, making the model hard to be trained. Considering that, we adopt an early-stop strategy to avoid gradient vanishing, which is formulated at Lines 7 to 9 in Algorithm~\ref{algo_solver}. Specifically, if the increment of the updated assignments is below a predefined threshold $\eta$, the updating of assignments and affinities will be stopped.
Furthermore, we set the iteration of the convolution operation in the AA-prediction network as 5 to capture the structural representations from 5-order neighborhoods. In the probabilistic solver, we specify the stop threshold $\eta = 0.00001$ and the iteration maximum $S = 10$ throughout our experiments.
During training, the positive label weight $w$ in Eq.~\ref{loss} is fixed to 5 to balance the significance of positive labels and negative labels.  Our model runs on a linux server with Intel(R) Xeon(R) CPU E5-2620 v4 (2.10GHz) and a TITAN XP GPU (12G).

\begin{table}[!t]
\centering
\caption{Comparison of matching accuracy (\%) on the Willow Object dataset. Bold \textbf{numbers} represent the best results.}
\label{table:willow}
\normalsize
\begin{tabular}
{@{\hspace{1.5mm}}l@{\hspace{1.mm}} | @{\hspace{1.mm}}c@{\hspace{1.mm}} @{\hspace{1.mm}}c@{\hspace{1.mm}} @{\hspace{1.mm}}c@{\hspace{1.mm}} @{\hspace{1.mm}}c@{\hspace{1.mm}} @{\hspace{1.mm}}c@{\hspace{1.mm}}
| @{\hspace{1.mm}}c@{\hspace{1.mm}}}
  \hline
    Algorithm & Car & Duck & Face & Motorb. & Wineb. & AVG\\
  \hline
%  IPFP~\cite{IPFP} &  74.8 &60.6 &98.9 &84.0 &79.0 &79.5  \\
%  RRWM~\cite{RWGM}&  86.3 &75.5 &\textcolor{red}{100} &94.9 &94.3 &90.2\\
%  PSM~\cite{PBGM}&    88.0 &76.8 &\textcolor{red}{100} &96.4 &97.0 &91.6 \\
%  GNCCP~\cite{GNCCP}& 86.4 &77.4 &\textcolor{red}{100} &95.6 &95.7 &91.0\\
%  ABPF~\cite{ABPF}&   88.4 &80.1 &\textcolor{red}{100} &96.2 &96.7 &92.3\\
%  \hline
%  HARG~\cite{LG2M}&  71.9 &72.2 &93.9 &71.4 &86.1 &79.1\\
  GMN~\cite{DGM}                & 74.3  & 82.8  & 99.3          & 71.4  &76.7   & 80.9 \\
  PCA~\cite{PCA}                & 84.0  & 93.5  & \textbf{100}  & 76.7  &96.9   & 90.2\\
  IPCA~\cite{PCA-PAMI}          & 90.2  & 84.9  & \textbf{100}  & 77.7  &95.2   & 89.6\\
  LGM~\cite{LGM}                & 91.2  & 86.2  & \textbf{100}  & 99.4  &97.9   & 94.9 \\
  qc-DGM~\cite{qcDGM}           & 98.0  & 92.8  & \textbf{100}  & 98.8  &\textbf{99.0}&97.7\\
  DGMC~\cite{DGM_consensus}     & 98.3  & 90.2  & \textbf{100}  & 98.5  & 98.1  & 97.0  \\
  CIE~\cite{Hungarian_Attention}& 82.2  & 81.2  & \textbf{100}  & 90.0  & 97.6  & 90.2  \\
  BBGM~\cite{BBGM}              & 96.9  & 89.0  & \textbf{100}  & 99.2  & 98.8  & 96.8  \\
  NGM~\cite{NGM}                & 97.4  &93.4   & \textbf{100}  & 98.6  & 98.3  & 97.5\\
  NGMv2~\cite{NGM}                & 97.4  &93.4   & \textbf{100}  & 98.6  & 98.3  & 97.5\\
  \hline
  DPGM (ours)                    &\textbf{99.5} & \textbf{96.8} & \textbf{100} & \textbf{100} & 93.0 & \textbf{97.9} \\
\hline
\end{tabular}
\end{table}

\begin{table*}[!htb]
%\small
\normalsize
\centering
\caption{Comparison of matching accuracy (\%) on the Pascal VOC dataset. Bold \textbf{numbers} represent the best results. }
\label{table:Pascal}
\scalebox{1.0}{
\begin{tabular}
{@{\hspace{.3mm}}c@{\hspace{.3mm}} | @{\hspace{.5mm}}c@{\hspace{.5mm}} @{\hspace{.5mm}}c@{\hspace{.5mm}} @{\hspace{.5mm}}c@{\hspace{.5mm}} @{\hspace{.5mm}}c@{\hspace{.5mm}} @{\hspace{.5mm}}c@{\hspace{.5mm}}
@{\hspace{.5mm}}c@{\hspace{.5mm}} @{\hspace{.5mm}}c@{\hspace{.5mm}} @{\hspace{.5mm}}c@{\hspace{.5mm}} @{\hspace{.5mm}}c@{\hspace{.5mm}} @{\hspace{.5mm}}c@{\hspace{.5mm}} @{\hspace{.5mm}}c@{\hspace{.5mm}} @{\hspace{.5mm}}c@{\hspace{.5mm}} @{\hspace{.5mm}}c@{\hspace{.5mm}} @{\hspace{.5mm}}c@{\hspace{.5mm}} @{\hspace{.5mm}}c@{\hspace{.5mm}} @{\hspace{.5mm}}c@{\hspace{.5mm}} @{\hspace{.5mm}}c@{\hspace{.5mm}} @{\hspace{.5mm}}c@{\hspace{.5mm}} @{\hspace{.5mm}}c@{\hspace{.5mm}} @{\hspace{.5mm}}c@{\hspace{.5mm}}
| @{\hspace{.3mm}}c@{\hspace{.3mm}}}
  \hline
    Algo. & aero & bike & bird & boat & bot. &bus & car &cat& cha.\ &cow & tab. &dog & hor.& mbi. & per. &pla. &she. &sofa& tra. & tv& AVG\\
  \hline
  %IPFP~\cite{IPFP} &  25.1 & 26.4 & 41.4 & 50.3 & 43.0 & 32.9 & 37.3 & 32.5 & 33.6 & 28.2 &  26.9 & 26.1 & 29.9 & 32.0 & 28.8 & 62.9 & 28.2 & 45.0 & 69.3 & 33.8 &36.6  \\
%  RRWM~\cite{RWGM}&  30.9 & 40.0 & 46.4 & 54.1 & 52.3 & 35.6 & 47.4 & 37.3 & 36.3 & 34.1 &28.8 & 35.0 & 39.1 & 36.2 & 39.5 &67.8 & 38.6 & 49.4 & 70.5& 41.3& 43.0  \\
%  PSM~\cite{PBGM}&     32.6 & 37.5 & 49.9 & 53.2 & 47.8 &34.6 & 50.1 & 35.5 & 37.2 & 36.3 &23.1 & 32.7 & 42.4 & 37.1 & 38.5 &62.3 & 41.7 & 54.3 & 72.6& 40.8 & 43.1 \\
%  GNCCP~\cite{GNCCP}&     28.9 & 37.1 & 46.2 & 53.1 & 48.0 & 36.3 & 45.5 & 34.7 & 36.3 & 34.2 &25.2 & 35.3 & 39.8 & 39.6 & 40.7 &61.9 & 37.4 & 50.5& 67.0 & 34.8 &41.6 \\
%  ABPF~\cite{ABPF}&     30.9 & 40.4 &47.3 &54.5 &50.8 &35.1 &46.7 &36.3 &40.9 &38.9 &16.3 &34.8& 39.8 &39.6 &39.3 &63.2 &37.9 &50.2 &70.5 &41.3&42.7 \\
%
%  \hline
  GMN~\cite{DGM} & 31.9 &47.2& 51.9& 40.8 &68.7 &72.2 &53.6 &52.8 &34.6 &48.6 &72.3 &47.7 &54.8 &51.0 &38.6 &75.1 &49.5 &45.0 &83.0 &86.3 &55.3  \\
  PCA~\cite{PCA}& 51.2 &61.3& 61.6 &58.4 &78.8 &73.9 &68.5 &71.1 &40.1 &63.3 &45.1 &64.4 &66.4 &62.2 &45.1 &79.1 &68.4 &60.0 &80.3 &91.9 &64.6 \\
  IPCA~\cite{PCA-PAMI} &51.0 &64.9 &68.4 &60.5 &80.2 &74.7 &71.0 &73.5 &42.2 &68.5 &48.9 &69.3 &67.6 &64.8 &48.6 &84.2 &69.8 &62.0 &79.3 &89.3 &66.9\\
  LGM~\cite{LGM}   & 46.9 &58.0& 63.6& 69.9& 87.8& 79.8& 71.8 &60.3 &44.8& 64.3 &79.4& 57.5& 64.4 &57.6 &52.4 &96.1& 62.9 &65.8 &94.4 &92.0& 68.5  \\
  qc-DGM~\cite{qcDGM}& 49.6 &64.6 &67.1& 62.4 &82.1& 79.9& 74.8& 73.5& 43.0& 68.4 &66.5 &67.2 &71.4 &70.1& 48.6 &92.4 &69.2 &70.9 &90.9& 92.0 &70.3\\
  DGMC~\cite{DGM_consensus} & 50.4 &67.6 &70.7 &70.5 &87.2& 85.2 &82.5 &74.3 &46.2 &69.4 &69.9 &73.9 &73.8& 65.4 &51.6 &98.0 &73.2& 69.6 &94.3 &89.6 &73.2  \\
  CIE~\cite{Hungarian_Attention} & 51.2 &69.2 &70.1 &55.0 &82.8& 72.8& 69.0& 74.2& 39.6 &68.8 &71.8 &70.0 &71.8 &66.8 &44.8 &85.2 &69.9& 65.4 &85.2 &92.4 &68.9  \\
  BBGM~\cite{BBGM} &61.5 &\textbf{75.0} & 78.1 & 80.0 & 87.4 & 93.0 & 89.1 & 80.2 &\textbf{58.1} &77.6 &76.5 &79.3 &78.6 &78.8 &66.7 &97.4 &76.4 &\textbf{77.5} &97.7 &\textbf{94.4} &80.1  \\
  NGM~\cite{NGM} &50.1 &63.5 &57.9 &53.4 &79.8 &77.1 &73.6& 68.2& 41.1& 66.4 &40.8 &60.3 &61.9& 63.5& 45.6 &77.1 &69.3 &65.5 &79.2 &88.2 &64.1 \\
  NGMv2~\cite{NGM}&61.8 &71.2 &77.6 &78.8 &87.3 &93.6 & 87.7 &79.8 &55.4 & 77.8 & 89.5 &78.8 & 80.1 &79.2 & 62.6 & 97.7 & 77.7 & 75.7 & 96.7 & 93.2 & 80.1\\
  \hline
  DPGM (ours) &\textbf{63.1} & 64.5 & \textbf{78.5} & \textbf{81.4} & \textbf{93.8} & 93.5 & \textbf{90.0} &\textbf{81.5} & 56.9 &\textbf{80.6} & \textbf{95.0} & \textbf{80.3} & \textbf{80.3} & 72.5 &\textbf{68.0} &\textbf{98.5} & \textbf{79.3} &75.4 & \textbf{98.3} & 92.8 &\textbf{81.2} \\
\hline
\end{tabular}
}
\end{table*}

\begin{table*}[!htb]
%\small
\normalsize
\centering
\caption{Comparison of matching accuracy (\%) on the Pascal VOC dataset with filtered image pairs. Bold \textbf{numbers} represent the best results. }
\label{table:filterVOC}
\scalebox{1.0}{
\begin{tabular}
{@{\hspace{.3mm}}c@{\hspace{.3mm}} | @{\hspace{.5mm}}c@{\hspace{.5mm}} @{\hspace{.5mm}}c@{\hspace{.5mm}} @{\hspace{.5mm}}c@{\hspace{.5mm}} @{\hspace{.5mm}}c@{\hspace{.5mm}} @{\hspace{.5mm}}c@{\hspace{.5mm}}
@{\hspace{.5mm}}c@{\hspace{.5mm}} @{\hspace{.5mm}}c@{\hspace{.5mm}} @{\hspace{.5mm}}c@{\hspace{.5mm}} @{\hspace{.5mm}}c@{\hspace{.5mm}} @{\hspace{.5mm}}c@{\hspace{.5mm}} @{\hspace{.5mm}}c@{\hspace{.5mm}} @{\hspace{.5mm}}c@{\hspace{.5mm}} @{\hspace{.5mm}}c@{\hspace{.5mm}} @{\hspace{.5mm}}c@{\hspace{.5mm}} | @{\hspace{.3mm}}c@{\hspace{.3mm}}}
  \hline
    Algo. & aero & bike & bird & boat & car &cat& cha. &cow &dog & hor.& mbi. & per. &she. &sofa& AVG\\
  \hline
  GMN~\cite{DGM} & 36.2 & 60.5 & 33.8 & 46.4 & 71.9 & 64.6 & 29.5 & 58.1 & 54.4 & 57.5 & 54.1 & 35.4 & 58.9 & \textbf{100.0} & 54.4  \\
  IPCA~\cite{PCA-PAMI}& 51.2 & 67.2 & 40.6 & 42.1 & 71.9 & 72.4 & 34.0 & 63.6 & 63.3 & 65.8 & 68.2 & 47.6 & 65.2 & \textbf{100.0} & 60.9  \\
  CIE~\cite{Hungarian_Attention} & 45.7 & 62.4 & 40.0 & 40.7 & 66.3 & 74.1 & 31.4 & 64.5 & 65.4 & 68.1 & 55.6 & 44.9 & 63.9 & 80.0 & 57.4\\
  BBGM~\cite{BBGM} &\textbf{57.0} & 72.6 & 50.0 & \textbf{60.7} & 88.6 & 79.7 & 44.8 & 75.8 & \textbf{76.9} & 76.5 & 71.5 & 64.6 & 71.2 & \textbf{100.0} & 70.7 \\
  NGMv2~\cite{NGM} &54.9 & \textbf{73.0} & 49.3 & 50.0 & 86.3 & 79.7 & \textbf{49.0} & \textbf{76.5} & 74.2 & 74.3 & \textbf{73.6} & 61.5 & 72.9 & \textbf{100.0} & 69.7 \\
  \hline
  DPGM (ours) &55.5 & 66.2 & \textbf{52.3} & 58.4 & \textbf{91.3} & \textbf{81.4} & 47.9 & 74.9 & 76.5 & \textbf{80.6} & 71.7 & \textbf{72.1} & \textbf{73.3} & \textbf{100.0} & \textbf{71.6}  \\
\hline
\end{tabular}
}
\end{table*}

\begin{table*}[!htb]
\normalsize
\centering
\caption{Comparison of matching accuracy (\%) on the SPair-71k dataset. Bold \textbf{numbers} represent the best results.}
\label{table:SPair71K}
\scalebox{1.0}{
\begin{tabular}
{@{\hspace{3.mm}}l@{\hspace{.3mm}} | @{\hspace{.5mm}}c@{\hspace{.5mm}} @{\hspace{.5mm}}c@{\hspace{.5mm}}  @{\hspace{.5mm}}c@{\hspace{.5mm}} @{\hspace{.5mm}}c@{\hspace{.5mm}} @{\hspace{.5mm}}c@{\hspace{.5mm}}
@{\hspace{.5mm}}c@{\hspace{.5mm}} @{\hspace{.5mm}}c@{\hspace{.5mm}} @{\hspace{.5mm}}c@{\hspace{.5mm}} @{\hspace{.5mm}}c@{\hspace{.5mm}} @{\hspace{.5mm}}c@{\hspace{.5mm}} @{\hspace{.5mm}}c@{\hspace{.5mm}} @{\hspace{.5mm}}c@{\hspace{.5mm}} @{\hspace{.5mm}}c@{\hspace{.5mm}} @{\hspace{.5mm}}c@{\hspace{.5mm}} @{\hspace{.5mm}}c@{\hspace{.5mm}} @{\hspace{.5mm}}c@{\hspace{.5mm}} @{\hspace{.5mm}}c@{\hspace{.5mm}} @{\hspace{.5mm}}c@{\hspace{.5mm}} | @{\hspace{.3mm}}c@{\hspace{.3mm}}}
  \hline
    Algo. & aero & bike & bird & boat & bott. &bus & car &cat& chair &cow &dog & hor.& mbi. & per. &plant &she. & train& tv& AVG\\
  \hline

  DGMC~\cite{DGM_consensus} & 54.8 &44.8 &80.3 &70.9 &65.5& 90.1 &78.5 &66.7 &66.4 &73.2 &66.2 &66.5 &65.7& 59.1 &98.7 &68.5 &84.9& 98.0 &72.2   \\
  BBGM~\cite{BBGM} &66.9 &57.7 &85.8& \textbf{78.5} &66.9 &95.4 &\textbf{86.1} &74.6 &\textbf{68.3} &78.9& \textbf{73.0}& 67.5& 79.3& 73.0& \textbf{99.1} &74.8 &\textbf{95.0} &98.6 &78.9 \\
  \hline
  DPGM (ours) & \textbf{68.5} &\textbf{64.0} &\textbf{86.6} &76.9 &\textbf{72.4} &\textbf{96.4}& 81.9& \textbf{75.9}& 65.6& \textbf{81.1}& \textbf{73.0}& \textbf{73.1}& \textbf{82.2}& \textbf{76.2}& 98.7& \textbf{83.1}& 89.0& \textbf{99.9}& \textbf{80.3} \\
\hline
\end{tabular}
}
\end{table*}
\subsection{Willow Object dataset}\label{sec:willow}
The Willow Object dataset provided in~\cite{LG2M} contains 5 categories, each of which is represented by at least 40 images with different instances. Each image in this dataset is annotated with 10 distinctive landmarks on the target object, which means there is no outlier keypoint between two images from the same category. Following the dataset split setting in~\cite{PCA,LGM}, we select 20 images from each category for training and keep the rest for testing. Note that any two images from the same class of the training set can form a training sample, and we thus obtain a total of 2,000 training samples. In testing, we randomly choose 1,000 pairs of images from the testing set of each category respectively.

The graph matching performances of the compared approaches are shown in Table~\ref{table:willow}. This dataset is considered relatively easy due to the lack of variation in pose, scale and illumination, thus most of the compared methods achieve excellent matching results, especially on the \emph{Face} category. By jointly learning pairwise affinities and initial node assignments, our method solves the quadratic assignment problem without  relaxation on the constraints in a deep probabilistic scheme. As a result, our method achieves the best matching accuracy in all categories expect \emph{Winebottle}, and surpasses all compared baselines in general.

%=====================================figure:bottle=================================
\begin{figure*}[!htb]
\centering
%\subfigure[image pair]
%{
%\includegraphics[width=2.65cm]{figs/pascal/bottle/image_pair.jpg}
%}
\subfigure[GMN: 0/8]{
\includegraphics[width=2.85cm]{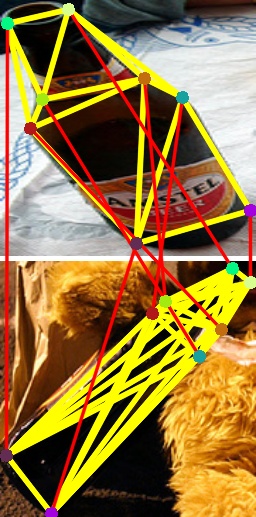}
}
\subfigure[PCA: 0/8]{
\includegraphics[width=2.85cm]{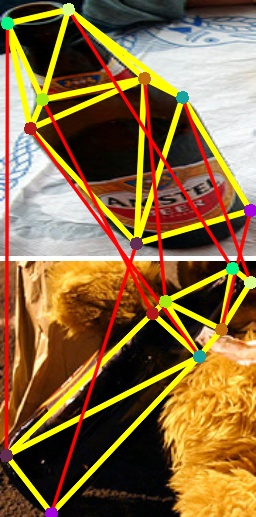}
%\caption{fig1}
}
\subfigure[IPCA: 2/8]{
\includegraphics[width=2.85cm]{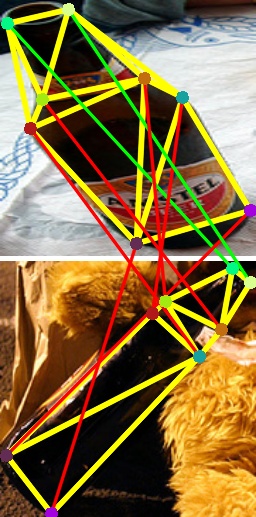}
}
\subfigure[CIE: 2/8]{
\includegraphics[width=2.85cm]{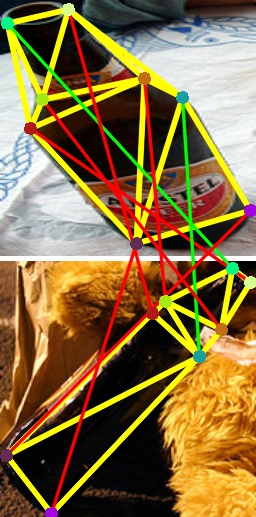}
}
\subfigure[DGMC: 3/8]{
\includegraphics[width=2.85cm]{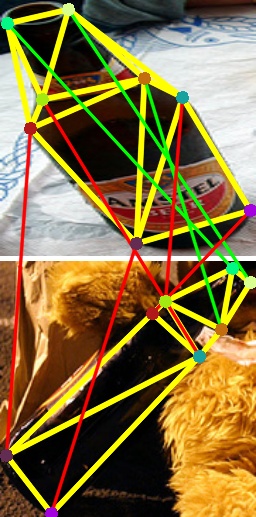}
}
\subfigure[LGM: 0/8]{
\includegraphics[width=2.85cm]{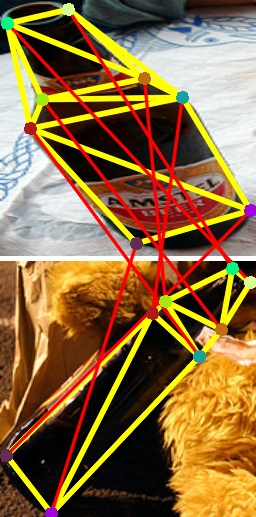}
}
\subfigure[NGM: 0/8]{
\includegraphics[width=2.85cm]{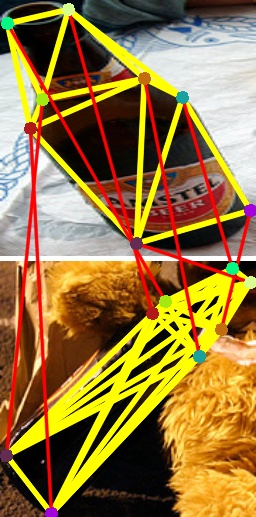}
}
\subfigure[NGMv2: 2/8]{
\includegraphics[width=2.85cm]{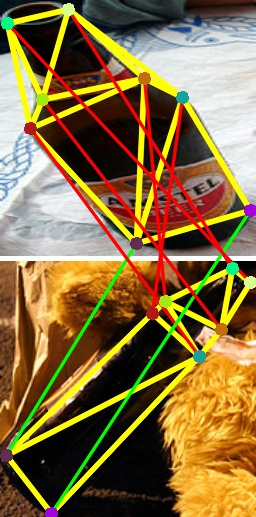}
}
\subfigure[BBGM: 2/8]{
\includegraphics[width=2.85cm]{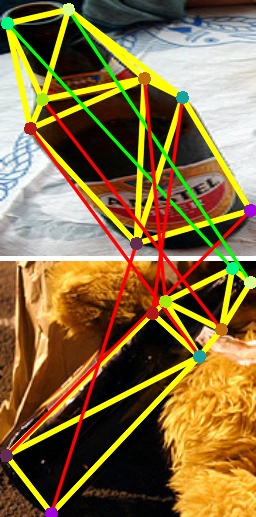}
}
%\subfigure[SuperGlue: 0/8]{
%\includegraphics[width=2.65cm]{figs/pascal/bottle/SG_H.jpg}
%}
\subfigure[DPGM(ours): 8/8]{
\includegraphics[width=2.85cm]{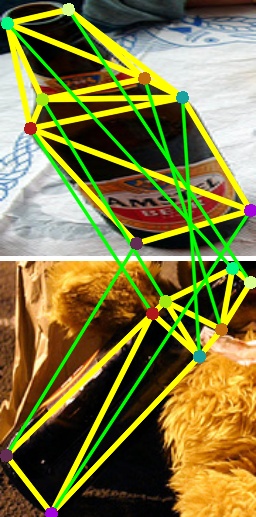}
}
\caption{The representative examples of matching results solved by all compared methods on category \emph{bottle}. The keypoints with same semantic annotations are marked in the same colors. The graph structures built in each image are represented as yellow lines. The incorrect and correct matches are represented as red lines and green lines across each pair of images, respectively.}
\label{fig:bottle}
\end{figure*}
%=====================================figure:bottle===================================

%=====================================figure:car=================================
\begin{figure*}[htbp]
\centering
%\subfigure[image pair]
%{
%\includegraphics[width=2.85cm]{figs/pascal/car/image_pair.jpg}
%}
\subfigure[GMN: 1/6]{
\includegraphics[width=2.85cm]{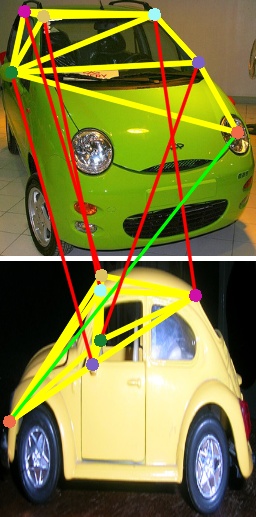}
}
\subfigure[PCA: 4/6]{
\includegraphics[width=2.85cm]{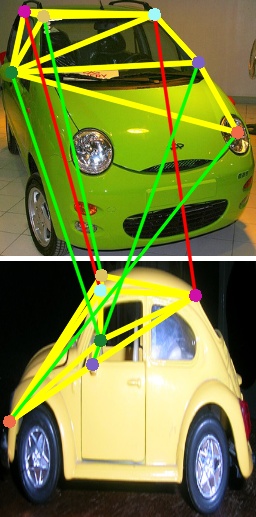}
}
\subfigure[IPCA: 2/6]{
\includegraphics[width=2.85cm]{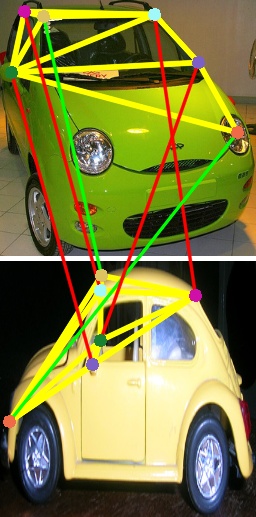}
}
\subfigure[CIE: 2/6]{
\includegraphics[width=2.85cm]{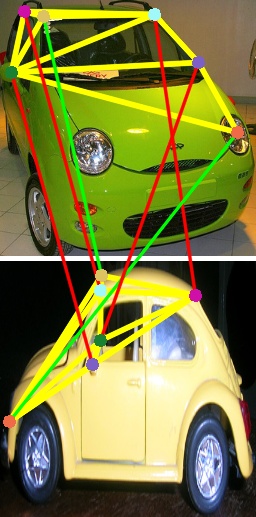}
}
\subfigure[DGMC: 0/6]{
\includegraphics[width=2.85cm]{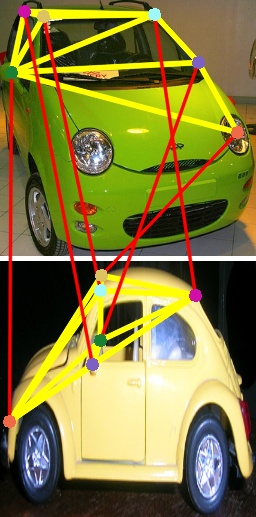}
}
\subfigure[LGM: 3/6]{
\includegraphics[width=2.85cm]{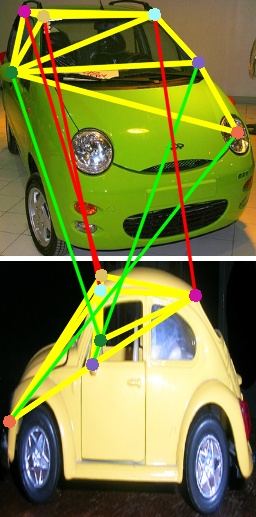}
}
\subfigure[NGM: 2/6]{
\includegraphics[width=2.85cm]{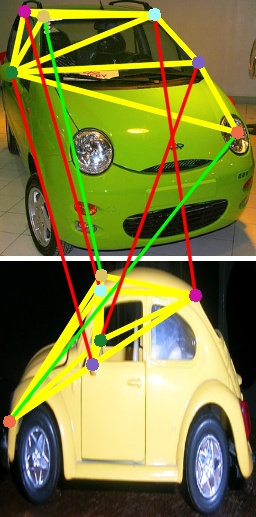}
}
\subfigure[NGMv2: 3/6]{
\includegraphics[width=2.85cm]{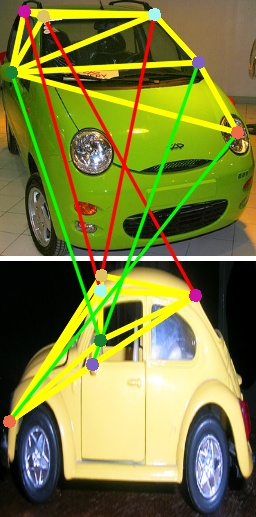}
}
\subfigure[BBGM: 2/6]{
\includegraphics[width=2.85cm]{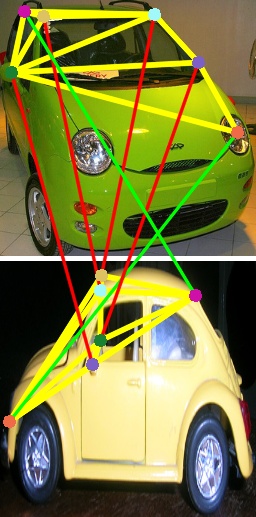}
}
%\subfigure[SuperGlue: 2/6]{
%\includegraphics[width=2.85cm]{figs/pascal/car/SG_H.jpg}
%}
\subfigure[DPGM(ours): 6/6]{
\includegraphics[width=2.85cm]{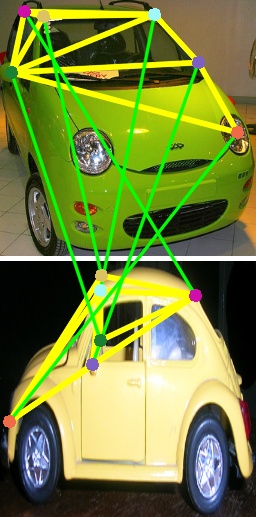}
}
\caption{The representative examples of matching solutions solved by all compared methods on category \emph{car}. The keypoints with same semantic annotations are marked in the same colors. The graph structures built in each image are represented as yellow lines. The incorrect and correct matches are represented as red lines and green lines across each pair of images, respectively.}
\label{fig:car}
\end{figure*}
%=====================================figure:car===================================
\subsection{Pascal VOC}
Pascal VOC~\cite{VOC} with Berkeley annotations of keypoints~\cite{berkeley} contains 20 classes of instances with labeled keypoint locations.
This dataset is considered more challenging than the Willow dataset for that instances may vary in its scale, pose, illumination and the number of inlier keypoints. Following previous arts~\cite{PCA,LGM}, we split the dataset into two groups, i.e., 7,020 annotated images for training and 1,682 images for testing. In particular, we randomly select 1,000 pairs of images from the testing set of each category for testing.

The comparison results on Pascal VOC~\cite{VOC} have been illustrated in Table~\ref{table:Pascal}.
Most of the previous methods~\cite{PCA,PCA-PAMI,LGM,qcDGM,DGM_consensus,Hungarian_Attention} that solve the graph matching problem in a relaxed form fail to provide satisfying matching results. After embedding a stronger combinatorial solver, BBGM~\cite{BBGM}, NGMv2~\cite{NGM} and our DPGM gain remarkable improvement in matching accuracy. In particular, our method achieves the best results on 14 categories and rises the average matching accuracy to $81.2\%$, surpassing the strongest competitors BBGM~\cite{BBGM} and NGMv2~\cite{NGM} by $1.1\%$. Some representative matching results of compared methods on categories~\emph{bottle} and~\emph{car} have been illustrated in Figs.~\ref{fig:bottle} and~\ref{fig:car}, respectively, both of which refer to the visual matching under heavy view changes. As illustrated in Figs.~\ref{fig:bottle} and~\ref{fig:car}, the previous deep learning methods, even those combined with combinatorial solvers, fail to obtain satisfied matching results. While with the proposed probabilistic solvers for searching the optimal solution from predicted assignments, our model DPGM achieves the best matching results where all the correct matches have been found.

Table~\ref{table:Pascal} illustrates the comparison of matching results on Pascal VOC~\cite{VOC} where each image pair may not contain the same keypoints, and all the compared methods only match an intersection of the two set of keypoints. However, this will lead to a problem that in many cases there may be only two 2 or 3 keypoints left. In such cases, what may contribute to the results is better visual feature, and the graph matching algorithm can not do much to improve the performance of the algorithm. Therefore, we perform an additional experiments on the Pascal VOC~\cite{VOC} dataset with filtered image pairs that contain at least 10 keypoints, for better validation of the contribution of graph matching algorithms. The comparison is reported in Table~\ref{table:filterVOC}\footnote{The parameters of the pretrained models in Table~\ref{table:filterVOC} are provided at: https://github.com/Thinklab-SJTU/ThinkMatch.} where only 14 categories of image pairs with at least 10 keypoints are kept.
As illustrated in Table~\ref{table:filterVOC}, all algorithms achieve relatively lower matching accuracies in comparison with that in the original Pascal VOC dataset (Table 2). This demonstrates that finding optimal matching between relatively large graphs is more difficult than that on small ones. While consistent with the results in Table~\ref{table:Pascal}, BBGM~\cite{BBGM} and NGMv2~\cite{NGM}, which embed combinatorial solvers into deep graph matching framework, achieve relatively higher average matching accuracy than other baselines. This again demonstrates that the combination of combinatorial solvers and deep affinity learning framework is more suitable for graph matching tasks. Furthermore, with the proposed probabilistic solver, our algorithm achieves comparable or even best matching accuracy on most categories in comparison with previous state-of-the-art baselines, and rises the average matching accuracy to 71.6\%, surpassing the strong competitors BBGM~\cite{BBGM} and NGMv2~\cite{NGM} by 0.9\% and 1.9\%, respectively, which validates the effectiveness of the proposed deep probabilistic solver.

\begin{table*}[!thb]
\normalsize
\centering
\caption{Comparison of inferring time (ms) on Willow and Pascal VOC.}
\label{table:time}
\scalebox{1.0}{
\begin{tabular}
{@{\hspace{.mm}}l@{\hspace{1.mm}} | @{\hspace{1.mm}}c@{\hspace{1.mm}} @{\hspace{1.mm}}c@{\hspace{1.mm}} @{\hspace{1.mm}}c@{\hspace{1.mm}} @{\hspace{1.mm}}c@{\hspace{1.mm}} @{\hspace{1.mm}}c@{\hspace{1.mm}}
@{\hspace{1.mm}}c@{\hspace{1.mm}} @{\hspace{1.mm}}c@{\hspace{1.mm}}@{\hspace{1.mm}}c@{\hspace{1.mm}}
@{\hspace{1.mm}}c@{\hspace{1.mm}}@{\hspace{1.mm}}c@{\hspace{1.mm}}}
  \hline
    Dataset & GMN & LGM & NGM & NGMv2 & PCA & IPCA & DGMC & CIE & BBGM & DPGM (ours)\\
  \hline
  Willow~\cite{LG2M}              & 28 & 42 & 14 & 15 & 29 & 35 & 3 & 28 & 13 & 10 \\
  Pascal VOC~\cite{VOC}           & 27 & 33 & 12 & 13 & 26 & 30 & 4 & 26 & 11 & 30\\
\hline
\end{tabular}
}
\end{table*}

\subsection{SPair-71k}
The SPair-71k~\cite{SPair} is another large-scale benchmark dataset including a total of 70,958 pairs of images from PASCAL 3D+~\cite{3DPascal} and Pascal VOC 2012~\cite{VOC2012}, which is very well organized with rich annotations for learning. Compared with Pascal VOC~\cite{VOC}, the pair annotations in this dataset have larger variations in view-point, scale, truncation and occlusion, thus reflecting the more generalized visual correspondence problem in real-world scenarios. In addition, it removes the ambiguous and poorly annotated categories, i.e., \emph{sofa} and \emph{dining table}, from the Pascal VOC dataset~\cite{VOC}.

The matching accuracy on SPair-71k~\cite{SPair} are reported in Table~\ref{table:SPair71K}, where two state-of-the-art algorithms DGMC~\cite{DGM_consensus} and BBGM~\cite{BBGM} are taken as the baselines in comparison with our method. Compared with DGMC~\cite{DGM_consensus}, the strongest competitor BBGM~\cite{BBGM} that employs a heavily optimized solver for graph matching achieves 6.7\% improvements on average matching accuracy, which demonstrates the effectiveness of the combination of deep framework for affinity learning and combinatorial solver. Furthermore, with the proposed probabilistic solver, our model DPGM achieves the best performance on most categories and reaches the best average matching accuracy of $80.3\%$, surpassing the DGMC~\cite{DGM_consensus} and BBGM~\cite{BBGM} by $8.1\%$ and $1.4\%$ respectively, which reveals that our method can be well generalized to more difficult graph matching problems in real-world scenarios.

\subsection{Running Time}
In addition to matching accuracy, computational efficiency is another important evaluation metric for graph matching methods. In this subsection, we report the inferring time of our algorithm in comparison with several deep-learning algorithms on the Pascal VOC~\cite{VOC} and Willow~\cite{LG2M} datasets. The comparisons for inferring time have been illustrated in Table~\ref{table:time}. It is observed that, our algorithm achieves the efficiency that is similar to or higher than GMN~\cite{DGM}, LGM~\cite{LGM}, NGM~\cite{NGM} and NGMv2~\cite{NGM} that are also based on Lawler' QAP formulation. On the other hand, all of them are in general comparable in computational efficiency with PCA~\cite{PCA}, IPCA~\cite{PCA-PAMI} and CIE~\cite{PCA}, which have relaxed the quadratic constraints. As the latent dimensions of the proposed model on Willow~\cite{LG2M} is relatively lower than that on Pascal VOC~\cite{VOC}, our model is more efficient for dealing with the visual matching tasks on Willow~\cite{LG2M} dataset. Specially, DGMC~\cite{DGM_consensus} is the most efficient one because the node features are complied in advance and the feature extraction module is removed from the code provided by the authors.

\subsection{Ablation Studies}

\begin{table*}[!thb]
\normalsize
\centering
\caption{Ablation studies on the Pascal VOC dataset. Bold \textbf{numbers} represent the best results.}
\label{table:ablation}
\scalebox{1.0}{
\begin{tabular}
{@{\hspace{.3mm}}c@{\hspace{.3mm}} | @{\hspace{.5mm}}c@{\hspace{.5mm}} @{\hspace{.5mm}}c@{\hspace{.5mm}} @{\hspace{.5mm}}c@{\hspace{.5mm}} @{\hspace{.5mm}}c@{\hspace{.5mm}} @{\hspace{.5mm}}c@{\hspace{.5mm}}
@{\hspace{.5mm}}c@{\hspace{.5mm}} @{\hspace{.5mm}}c@{\hspace{.5mm}} @{\hspace{.5mm}}c@{\hspace{.5mm}} @{\hspace{.5mm}}c@{\hspace{.5mm}} @{\hspace{.5mm}}c@{\hspace{.5mm}} @{\hspace{.5mm}}c@{\hspace{.5mm}} @{\hspace{.5mm}}c@{\hspace{.5mm}} @{\hspace{.5mm}}c@{\hspace{.5mm}} @{\hspace{.5mm}}c@{\hspace{.5mm}} @{\hspace{.5mm}}c@{\hspace{.5mm}} @{\hspace{.5mm}}c@{\hspace{.5mm}} @{\hspace{.5mm}}c@{\hspace{.5mm}} @{\hspace{.5mm}}c@{\hspace{.5mm}} @{\hspace{.5mm}}c@{\hspace{.5mm}} @{\hspace{.5mm}}c@{\hspace{.5mm}}
| @{\hspace{.3mm}}c@{\hspace{.3mm}}}
  \hline
    Abl.  & aero & bike & bird & boat & bott. &bus & car &cat& chair &cow & tab. &dog & hor.& mbi. & per. &plant &she. &sofa& train& tv& AVG\\
  \hline
  WPS  &56.4 & 65.8&74.7&78.0&88.2&92.5&85.6&77.6&54.9&73.9&83.5&74.8&75.0&71.5&61.7&95.7&75.5&74.4&96.6&92.4&77.4\\
  TIA  &56.7 & \textbf{69.4} & 75.4 & 74.8 & 91.5 & 87.4 & 87.0 & 76.4 & 52.2 & 76.2 & 80.1 & 77.3 & 76.5 & 71.9 & 60.4 & 97.5 & 75.4 &67.4 &96.7 &87.7 &76.9\\
  \hline
  DPGM &\textbf{63.1} & 64.5 & \textbf{78.5} & \textbf{81.4} & \textbf{93.8} &\textbf{93.5} & \textbf{90.0} &\textbf{81.5} & \textbf{56.9} &\textbf{80.6} &\textbf{95.0} & \textbf{80.3} & \textbf{80.3} & \textbf{72.5} &\textbf{68.0} &\textbf{98.5} & \textbf{79.3} &\textbf{75.4}&\textbf{98.3} & \textbf{92.8}&\textbf{81.2} \\
\hline
\end{tabular}
}
\end{table*}

\begin{table*}[!thb]
\normalsize
\centering
\caption{Ablation studies on the SPair-71k dataset. Bold \textbf{numbers} represent the best results.}
\label{table:abl_spair}
\scalebox{1.0}{
\begin{tabular}
{@{\hspace{3.mm}}l@{\hspace{.3mm}} | @{\hspace{.5mm}}c@{\hspace{.5mm}} @{\hspace{.5mm}}c@{\hspace{.5mm}}  @{\hspace{.5mm}}c@{\hspace{.5mm}} @{\hspace{.5mm}}c@{\hspace{.5mm}} @{\hspace{.5mm}}c@{\hspace{.5mm}}
@{\hspace{.5mm}}c@{\hspace{.5mm}} @{\hspace{.5mm}}c@{\hspace{.5mm}} @{\hspace{.5mm}}c@{\hspace{.5mm}} @{\hspace{.5mm}}c@{\hspace{.5mm}} @{\hspace{.5mm}}c@{\hspace{.5mm}} @{\hspace{.5mm}}c@{\hspace{.5mm}} @{\hspace{.5mm}}c@{\hspace{.5mm}} @{\hspace{.5mm}}c@{\hspace{.5mm}} @{\hspace{.5mm}}c@{\hspace{.5mm}} @{\hspace{.5mm}}c@{\hspace{.5mm}} @{\hspace{.5mm}}c@{\hspace{.5mm}} @{\hspace{.5mm}}c@{\hspace{.5mm}} @{\hspace{.5mm}}c@{\hspace{.5mm}} | @{\hspace{.3mm}}c@{\hspace{.3mm}}}
  \hline
    Abl. & aero & bike & bird & boat & bott. &bus & car &cat& chair &cow &dog & hor.& mbi. & per. &plant &she. & train& tv& AVG\\
  \hline

  WPS  & 65.3 & 56.1& 84.4 &75.0 &\textbf{72.7}& \textbf{96.9} &\textbf{85.6}& 74.0 &64.0 &\textbf{81.5}& 70.4& 68.7 &72.8 &72.7& \textbf{99.0}& 78.9 &\textbf{93.3}& 99.8 &78.4   \\
  TIA & 68.2& 52.1& 84.3& 73.9&70.8&95.8&82.8&\textbf{76.2}&60.0&78.4&71.2&67.0&69.5&72.4&96.6&78.5&92.1&99.7&77.2 \\
  \hline
  DPGM & \textbf{68.5} &\textbf{64.0} &\textbf{86.6} &\textbf{76.9} &72.4 &96.4& 81.9& 75.9& \textbf{65.6}& 81.1& \textbf{73.0}& \textbf{73.1}& \textbf{82.2}& \textbf{76.2}& 98.7& \textbf{83.1}& 89.0& \textbf{99.9}& \textbf{80.3} \\
\hline
\end{tabular}
}
\end{table*}

\begin{table}[!thb]
\normalsize
\centering
\caption{Ablation studies on the Willow Object dataset. Bold \textbf{numbers} represent the best results.}
\label{table:abl_willow}
\scalebox{1.0}{
\begin{tabular}
{@{\hspace{.mm}}l@{\hspace{1.mm}} | @{\hspace{1.mm}}c@{\hspace{1.mm}} @{\hspace{1.mm}}c@{\hspace{1.mm}} @{\hspace{1.mm}}c@{\hspace{1.mm}} @{\hspace{1.mm}}c@{\hspace{1.mm}} @{\hspace{1.mm}}c@{\hspace{1.mm}}
| @{\hspace{1.mm}}c@{\hspace{1.mm}}}
  \hline
    Ablation & Car & Duck & Face & Motorb. & Wineb. & AVG\\
  \hline
  WPS              & 91.8  & 85.4  & \textbf{100}  & 98.3  & 97.1  & 95.4  \\
  TIA                & 96.7  &86.3   & \textbf{100}  & 98.2  & \textbf{99.0}  & 96.0\\
  \hline
  DPGM                   &\textbf{99.5} & \textbf{96.8} & \textbf{100} & \textbf{100} & 93.0 & \textbf{97.9} \\
\hline
\end{tabular}
}
\end{table}

\subsubsection{Contributions of learnt assignments and affinities}
For illustration of the contributions of different modules in our framework, we provide various ablation studies and report results of two downgraded versions of our algorithm, named TIA and WPS, in Tables~\ref{table:ablation},~\ref{table:abl_spair},~\ref{table:abl_willow}.

\textbf{Trivial initialization of assignments (TIA):} we exclude the estimated assignments from the output of the AA-predictor, and initialize trivial assignments as the input of the probabilistic solver.

\textbf{Without probabilistic solver (WPS):} the differentiable probabilistic solver is removed from the network, and the output assignments of the AA-decoder are directly taken as the final solutions.

As shown in Tables~\ref{table:ablation}, \ref{table:abl_spair}, \ref{table:abl_willow}, both the downgraded versions of our model obtain
inferior performance to the full version on all the three benchmarks. Especially for Pascal VOC~\cite{VOC}, the full version of our model outperforms TIA and WPS by 4.3\% and 3.8\% on the average matching accuracy, respectively. These studies show that both the components make contributions to the excellent performance of the proposed DPGM algorithm.

Furthermore, we also explore the studies that combine the learnt affinities and other traditional learning-free solvers on Pascal VOC~\cite{VOC}. Specifically, we take the affinities learnt by our AA-prediction network as input, and adopt several state-of-the-art traditional learning-free solvers, including RRWM~\cite{RWGM}, PSM~\cite{PBGM} and IPFP~\cite{IPFP}, to search the optimal solutions. The comparisons on Pascal VOC~\cite{VOC} are reported in Table~\ref{table:HL_K}, where RRWM-H and RRWM-L denote the results of using the handcrafted affinities and the learned ones, respectively. Same notations are used for PSM~\cite{PBGM} and IPFP~\cite{IPFP}.  The results of using the handcrafted affinities are cited from LGM~\cite{LGM}. Obviously, replacing the handcrafted affinities with the learned ones can significantly improve the matching accuracies in most categories, and rise the average matching accuracies by 35.6\%, 37.3\% and 39.2\% for RRWM~\cite{RWGM}, PSM~\cite{PBGM} and IPFP~\cite{IPFP}, respectively. It clearly demonstrates the effectiveness of the learned affinities by our framework for graph matching. On the other hand, even though the learnt affinities are combined, these traditional solvers are still inferior to our framework by a certain gaps. It indicates that the learnt affinities are more suitable for end-to-end learnable framework than the transitional learning-free solvers.

%=====================================figure:ablation=================================
\begin{figure}[!htb]
\centering
\subfigure[]{
\label{fig:AA}
\includegraphics[width=8.5cm]{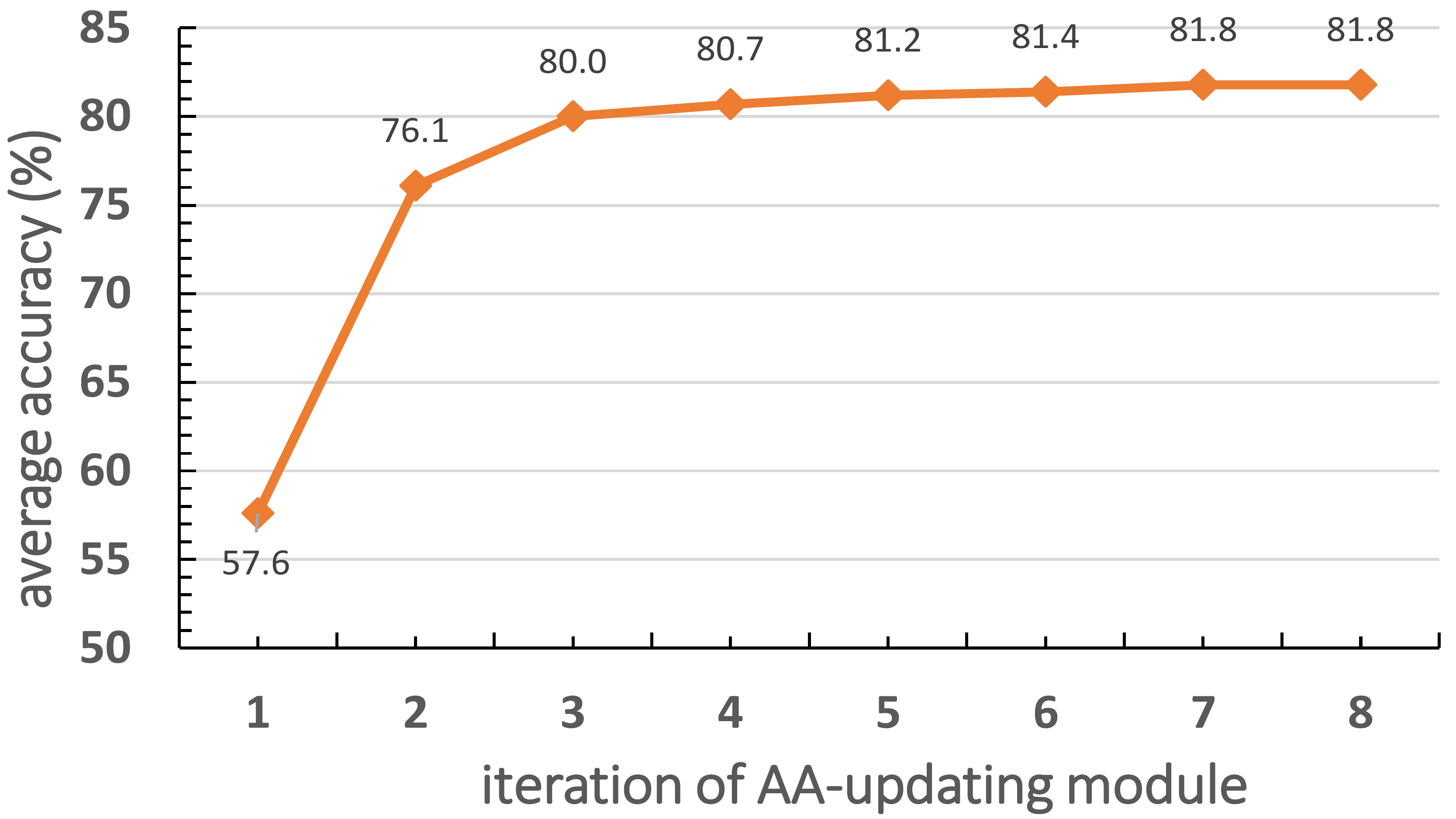}
}
\subfigure[]{
\label{fig:S}
\includegraphics[width=8.5cm]{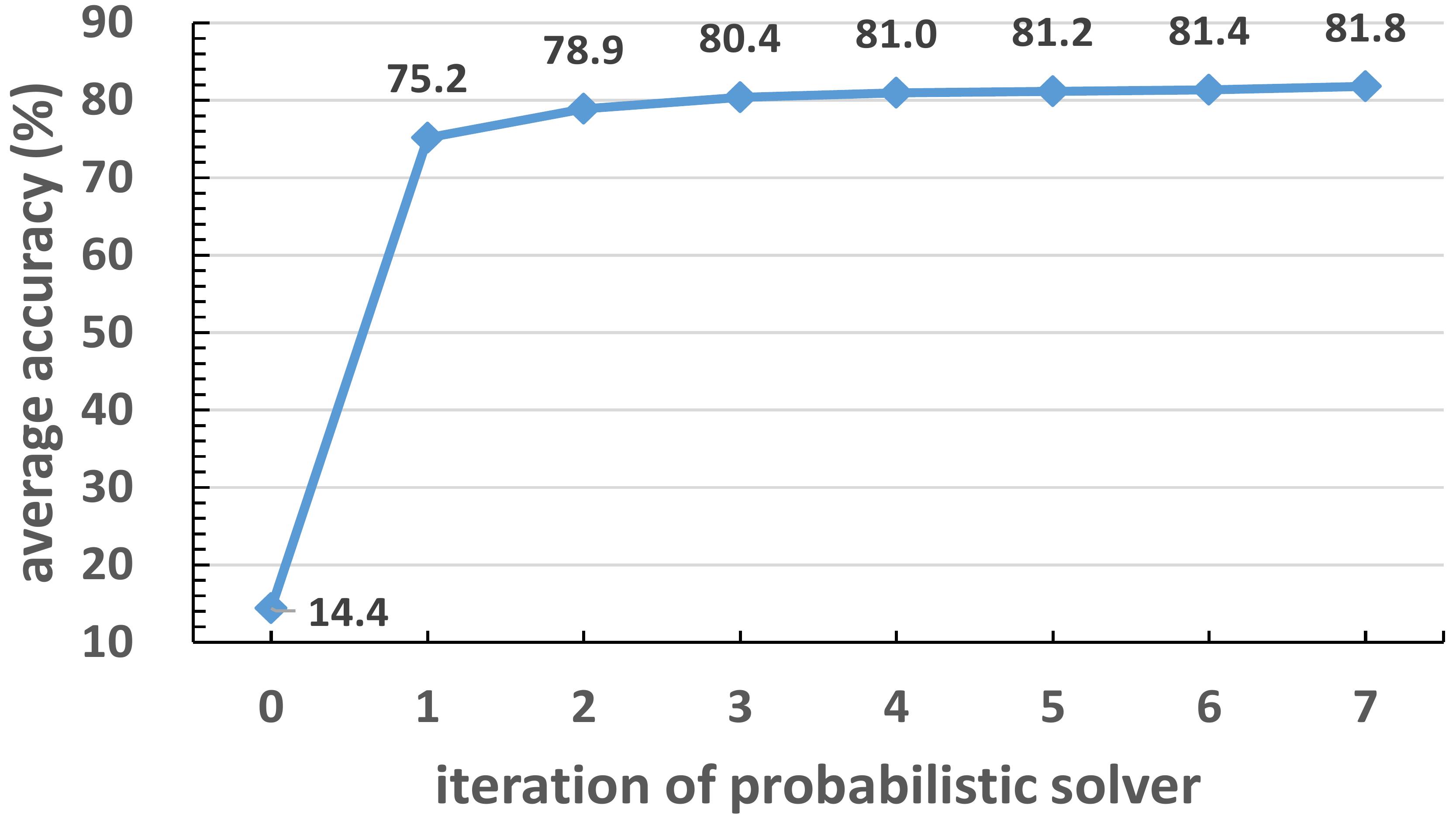}
}

\caption{Results in the ablation experiments on the number of iterations for (a) the AA-updating module and (b) the probabilistic solver.}
\end{figure}

\begin{table*}[!thb]
\normalsize
\centering
\caption{Comparison of matching accuracy (\%) on the Pascal VOC dataset under the combination of learnt affinities and traditional learning-free solvers. Bold \textbf{numbers} represent the best results. }
\label{table:HL_K}
\scalebox{1.0}{
\begin{tabular}
{@{\hspace{.3mm}}c@{\hspace{.3mm}} | @{\hspace{.5mm}}c@{\hspace{.5mm}} @{\hspace{.5mm}}c@{\hspace{.5mm}} @{\hspace{.5mm}}c@{\hspace{.5mm}} @{\hspace{.5mm}}c@{\hspace{.5mm}} @{\hspace{.5mm}}c@{\hspace{.5mm}}
@{\hspace{.5mm}}c@{\hspace{.5mm}} @{\hspace{.5mm}}c@{\hspace{.5mm}} @{\hspace{.5mm}}c@{\hspace{.5mm}} @{\hspace{.5mm}}c@{\hspace{.5mm}} @{\hspace{.5mm}}c@{\hspace{.5mm}} @{\hspace{.5mm}}c@{\hspace{.5mm}} @{\hspace{.5mm}}c@{\hspace{.5mm}} @{\hspace{.5mm}}c@{\hspace{.5mm}} @{\hspace{.5mm}}c@{\hspace{.5mm}} @{\hspace{.5mm}}c@{\hspace{.5mm}} @{\hspace{.5mm}}c@{\hspace{.5mm}} @{\hspace{.5mm}}c@{\hspace{.5mm}} @{\hspace{.5mm}}c@{\hspace{.5mm}} @{\hspace{.5mm}}c@{\hspace{.5mm}} @{\hspace{.5mm}}c@{\hspace{.5mm}}
| @{\hspace{.3mm}}c@{\hspace{.3mm}}}
  \hline
    Algo. & aero & bike & bird & boat & bot. &bus & car &cat& cha.\ &cow & tab. &dog & hor.& mbi. & per. &pla. &she. &sofa& tra. & tv& AVG\\
  \hline
  RRWM-H&  30.9 & 40.0 & 46.4 & 54.1 & 52.3 & 35.6 & 47.4 & 37.3 & 36.3 & 34.1 &28.8 & 35.0 & 39.1 & 36.2 & 39.5 &67.8 & 38.6 & 49.4 & 70.5& 41.3& 43.0  \\
  %RRWM-L & 37.0 & 32.8 & 49.6 & 74.6 & 80.9 & 84.0 & 71.9 & 39.7 & 34.0 & 46.5 & 51.7 & 43.2 & 45.9 & 40.4 & 52.0 & 94.5 & 46.6 & 60.2 & 89.9 & 58.1 & 56.7 \\
  RRWM-L & 61.5 & 61.0 & 76.8 & 79.8 & 93.1 & 91.4 & 88.5 & 78.2 & 54.2 & 76.8 & 98.2 & 75.5 & 76.0 & 69.5 & 65.5 & 98.1 & 75.8 & 68.9 & 88.6 & 93.6 & 78.6\\

  PSM-H&     32.6 & 37.5 & 49.9 & 53.2 & 47.8 &34.6 & 50.1 & 35.5 & 37.2 & 36.3 &23.1 & 32.7 & 42.4 & 37.1 & 38.5 &62.3 & 41.7 & 54.3 & 72.6& 40.8 & 43.1 \\
  %PSM-L& 37.9 & 32.5 & 47.2 & 75.7 & 82.7 & 86.6 & 73.1 & 37.0 & 33.4 & 48.9 & 65.7 & 41.6 & 46.1 & 39.0 & 54.8 & 96.3 & 47.3 & 61.3 & 83.6 & 49.4 & 57.0 \\
  PSM-L& 62.4 & 64.3 & 78.0 & 81.3 & 93.2 & 93.0 & 89.9 & 81.3 & 55.6 & 79.9 & \textbf{100.0} & 78.0 & 80.0 & 72.2 & 66.8 & 98.3 & 77.2 & 73.2 & 90.6 & \textbf{93.9} & 80.4\\

  IPFP-H &  25.1 & 26.4 & 41.4 & 50.3 & 43.0 & 32.9 & 37.3 & 32.5 & 33.6 & 28.2 &  26.9 & 26.1 & 29.9 & 32.0 & 28.8 & 62.9 & 28.2 & 45.0 & 69.3 & 33.8 &36.6  \\
  %IPFP-L & 36.3 & 30.8 & 50.6 & 72.6 & 76.4 & 74.0 & 69.2 & 40.7 & 31.0 & 43.7 & 52.8 & 43.6 & 42.6 & 41.3 & 46.2 & 93.2 & 45.9 & 57.1 & 88.7 & 57.5 & 54.7 \\
  IPFP-L & 59.5 & 59.1 & 74.2 & 78.1 & 91.4 & 87.7 & 85.7 & 76.4 & 52.3 & 73.8 & 84.7 & 73.8 & 73.9 & 66.7 & 62.8 & 97.8 & 73.9 & 67.6 & 84.1 & 93.6 & 75.8\\

  \hline
  DPGM &\textbf{63.1} & \textbf{64.5} & \textbf{78.5} & \textbf{81.4} & \textbf{93.8} & \textbf{93.5} & \textbf{90.0} &\textbf{81.5} &\textbf{ 56.9} &\textbf{80.6} & 95.0 & \textbf{80.3} & \textbf{80.3} & \textbf{72.5} &\textbf{68.0} &\textbf{98.5} & \textbf{79.3} &\textbf{75.4} & \textbf{98.3} & 92.8 &\textbf{81.2} \\
\hline
\end{tabular}
}
\end{table*}

%=====================================figure:ablation===================================
\subsubsection{Iteration of AA-updating module}
In the AA-predictor, the iteration of AA-updating module determines the receptive field of aggregated information from neighbors, which has direct influence to the final matching performance. Here, we vary the AA-updating iteration from 1 to 8 to explore its influence to the average matching accuracy. As illustrated in Fig.~\ref{fig:AA}, the average matching accuracy is only 57.6\% when only single iteration of AA-updating is performed. While the performance of our model can be significantly improved by setting the iteration of AA-updating from 2 to 5. However, there is no evident improvement when the iteration of AA-updating module is set as a value greater than 5, which demonstrates only 5 iterations of AA-updating module is sufficient to capture the structural representations of AA-graph. This in turn explain why we set the convolution iteration in AA-predictor as 5.

\subsubsection{Iteration of probabilistic solver}
Given the predicted assignments and affinities by AA-predictor, the proposed probabilistic solver refines them iteratively to search the optimal solution. To explore the influence of iteration of probabilistic solver to the model performance, we vary the refinement iteration for assignments from 0 to 7. As illustrated in Fig.~\ref{fig:S}, without refinement for assignments, i.e., setting the refinement iteration as 0, the average matching accuracy is only 14.4\%. With a single iteration of refinement for assignments, our model rises the matching accuracy to 75.2\%, which surpasses most of the state-of-the-art baselines. When setting the iteration of probabilistic solver as a value greater than 4, the average matching accuracy is stable around 81.0\%. The illustration of results in Fig.~\ref{fig:S} and the convergence showed in Fig~\ref{subfig:sparse} reveal that our model needs not many refinement iteration to search the optimal solutions, which guarantees the efficiency of the proposed model.

\section{Conclusion}
In this paper, we propose a novel deep learning algorithm, named DPGM, for graph matching.
In DPGM, an affinity-assignment prediction network is developed to learn the pairwise affinities and at the same time estimate the initial node assignments, and a differentiable solver is embedded to better optimize the QAP in a probabilistic perspective without compromise on the matching constraints.
Experimental results reveal that DPGM achieves state-of-the-art matching performance on various real-world image datasets.

%In addition, the ablation studies demonstrate that both the learnt pairwise affinities and the predicted assignments by our AA-predictor are beneficial to the final solution. Furthermore, combined with learnt affinities, several traditional learning-free graph matching methods achieves significant improvement on graph matching accuracy, which reveal that our AA-predictor can well capture the reasonable affinities between pair of candidate matches.

\section*{Acknowledgment}
\noindent This work is supported by the National Nature Science Foundation of China (Nos. 62076021, 62072027 and 61872032) and the Beijing Municipal Natural Science Foundation (Nos. 4202060, 4202057 and 4212041).

\bibliographystyle{IEEEtran}
\bibliography{IEEEabrv,reference}

\begin{IEEEbiography}[{\includegraphics[width=1in,height=1.25in,clip,keepaspectratio]{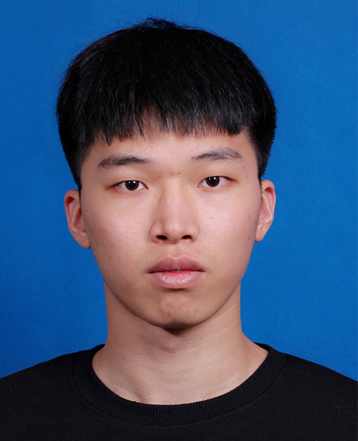}}]{He Liu}
received the B. S. degree in compute science from Hebei University of Technology, Tianjin, China. He is currently pursuing the Ph. D degree in the School of Computer and Information Technology, Beijing Jiaotong University, Beijing, China. His research concentrates on machine learning and graph matching.
\end{IEEEbiography}

\begin{IEEEbiography}[{\includegraphics[width=1in,height=1.25in,clip,keepaspectratio]{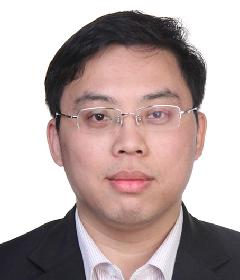}}]{Tao Wang}
received the PhD degree from the School of Computer and Information Technology, Beijing Jiaotong University, Beijing, China, in 2013. He was a visiting professor with the Department of Computer and Information Sciences, Temple University, from 2014 to 2015. He is currently an associate professor with the School of Computer and Information Technology, Beijing Jiaotong University. His research interests include graph matching theory with application to image analysis and retrieval.
\end{IEEEbiography}

\begin{IEEEbiography}[{\includegraphics[width=1in,height=1.25in,clip,keepaspectratio]{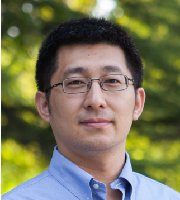}}]{Yidong Li}
was born in 1982, native to Shanxi, associate professor and Ph.D. supervisor. In 2003, he graduated from the Department of information and communication engineering of Beijing Jiaotong University. He received master and Ph. D. degree from the department of computer science of the University of Adelaide in Australia in 2006 and 2011 respectively. Dr. Yi is the vice dean of the school of computer and information technology, Beijing Jiaotong University, executive director of the SAP University Competence Center (China).
\end{IEEEbiography}

\begin{IEEEbiography}[{\includegraphics[width=1in,height=1.25in,clip,keepaspectratio]{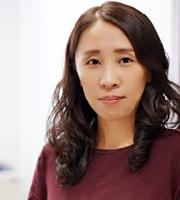}}]{Congyan Lang}
received her Ph.D. degree from the School of Computer and Information Technology, Beijing Jiaotong University, Beijing, China, in 2006. She was a Visiting Professor with the Department of Electrical and Computer Engineering, National
University of Singapore, Singapore, from 2010 to 2011. From 2014 to 2015, she visited the Department of Computer Science, University of Rochester, Rochester, NY, USA, as a Visiting Professor. She is currently a Professor with the School of Computer
and Information Technology, Beijing Jiaotong University. Her research areas include computer vision, machine learning, object recognition and segmentation.
\end{IEEEbiography}

\begin{IEEEbiography}[{\includegraphics[width=1in,height=1.25in,clip,keepaspectratio]{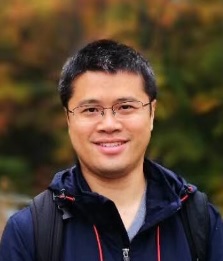}}]{Songhe Feng}
received the Ph.D. degree from the School of Computer and Information Technology, Beijing Jiaotong University, Beijing, China, in 2009. He was a Visiting Scholar with the Department of Computer Science and Engineering, Michigan State University, USA, from 2013 to 2014. He is currently a Professor with the School of Computer and Information Technology, Beijing Jiaotong University. His research interests include computer vision and machine learning.
\end{IEEEbiography}

\begin{IEEEbiography}[{\includegraphics[width=1in,height=1.25in,clip,keepaspectratio]{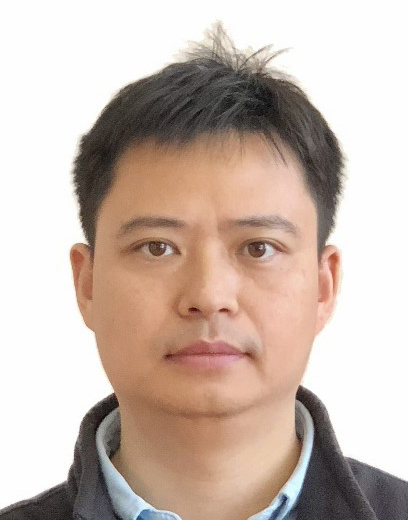}}] {Haibin Ling} received Ph.D. degree from University of Maryland in 2006. From 2007-2008, he worked for Siemens Corporate Research as a research scientist; and from 2008 to 2019, he was a faculty member of the Department of Computer Sciences for Temple University. In fall 2019, he joined the Department of Computer Science of Stony Brook University, where he is now a SUNY Empire Innovation Professor. His research interests include computer vision, augmented reality, medical image analysis, visual privacy protection, and human computer interaction.
\end{IEEEbiography}

\end{document}